\documentclass[journal]{IEEEtran}
\usepackage[numbers,sort&compress]{natbib}
\usepackage{booktabs}

\usepackage{multirow}
\usepackage{amssymb}
\usepackage[table ]{ xcolor}
\usepackage{threeparttable}
\usepackage{float}
\usepackage[misc]{ifsym}
\usepackage{makecell}

\ifCLASSINFOpdf
  \usepackage{amsmath}
  \usepackage[pdftex]{graphicx}
  \graphicspath{{../pdf/}{../jpg/}}
  \DeclareGraphicsExtensions{.pdf,.jpg,.png}
\else
  \usepackage[dvips]{graphicx}
  \graphicspath{{../eps/}}
\fi
\begin{document}
%

\title{Bare Demo of IEEEtran.cls\\ for IEEE Journals}
%$^{(\textrm{\Letter})}$

\author{Nana~Yu, Hong~Shi and Yahong~Han % <-this % stops a space
\thanks{This work is supported by the CAAI-Huawei MindSpore Open Fund and NSFC (under Grant 62376186, 61932009). Nana Yu, Hong Shi and Yahong Han are with the College of Intelligence and Computing, Tianjin Key Lab of Machine Learning, Tianjin University, Tianjin 300072, China. E-mail: \{yunana, serena, yahong\}@tju.edu.cn. }
%\thanks{Hong Shi is with the College of Intelligence and Computing, Tianjin Key Lab of Machine Learning, Tianjin University, Tianjin 300072, China. E-mail: serena@tju.edu.cn}
%\thanks{Yahong Han is with the College of Intelligence and Computing, Tianjin Key Lab of Machine Learning, Tianjin University, Tianjin 300072, China. E-mail: yahong@tju.edu.cn}
%\thanks{L. Fan is with School of Computer Science and Technology, Shandong University of Finance and Economics, Jinan, 250014, China (e-mail: fanlinwei@sdufe.edu.cn)}
}
% <-this % stops a space

\markboth{Journal of \LaTeX\ Class Files,~Vol.~14, No.~8, August~2015}%
{Shell \MakeLowercase{\textit{et al.}}: Bare Demo of IEEEtran.cls for IEEE Journals}



\title{Joint Correcting and Refinement for Balanced Low-Light Image Enhancement}
%\title{A Balanced Approach for Joint Correction and Refinement in Low-Light Image Enhancement}

\maketitle

\begin{abstract}
Low-light image enhancement tasks demand an appropriate balance among brightness, color, and illumination. While existing methods often focus on one aspect of the image without considering how to pay attention to this balance, which will cause problems of color distortion and overexposure etc. This seriously affects both human visual perception and the performance of high-level visual models. In this work, a novel synergistic structure is proposed which can balance brightness, color, and illumination more effectively. Specifically, the proposed method, so-called Joint Correcting and Refinement Network (JCRNet), which mainly consists of three stages to balance brightness, color, and illumination of enhancement. Stage 1: we utilize a basic encoder-decoder and local supervision mechanism to extract local information and more comprehensive details for enhancement. Stage 2: cross-stage feature transmission and spatial feature transformation further facilitate color correction and feature refinement. Stage 3: we employ a dynamic illumination adjustment approach to embed residuals between predicted and ground truth images into the model, adaptively adjusting illumination balance. Extensive experiments demonstrate that the proposed method exhibits comprehensive performance advantages over 21 state-of-the-art methods on 9 benchmark datasets. Furthermore, a more persuasive experiment has been conducted to validate our approach the effectiveness in downstream visual tasks (e.g., saliency detection). Compared to several enhancement models, the proposed method effectively improves the segmentation results and quantitative metrics of saliency detection. The source code will be available at https://github.com/woshiyll/JCRNet.

\end{abstract}


% \fi
\IEEEpeerreviewmaketitle

\section{Introduction}
\IEEEPARstart{A}{s} an important pre-processing stage, low-light image enhancement not only improves the visual experience of users but also enhances the feasibility of image analysis and understanding tasks \cite{hoyer2022daformer, li2022deep, ji2022asymmetric, han2021visual}. This is beneficial for the design and implementation of hardware and software in autonomous driving, security, and other systems. Recently, progress has been made in low-light image enhancement, such as contrast enhancement and noise reduction, both traditionally and with deep learning.
\begin{figure}[!t]
	\centering
	\includegraphics[width=0.5\textwidth]{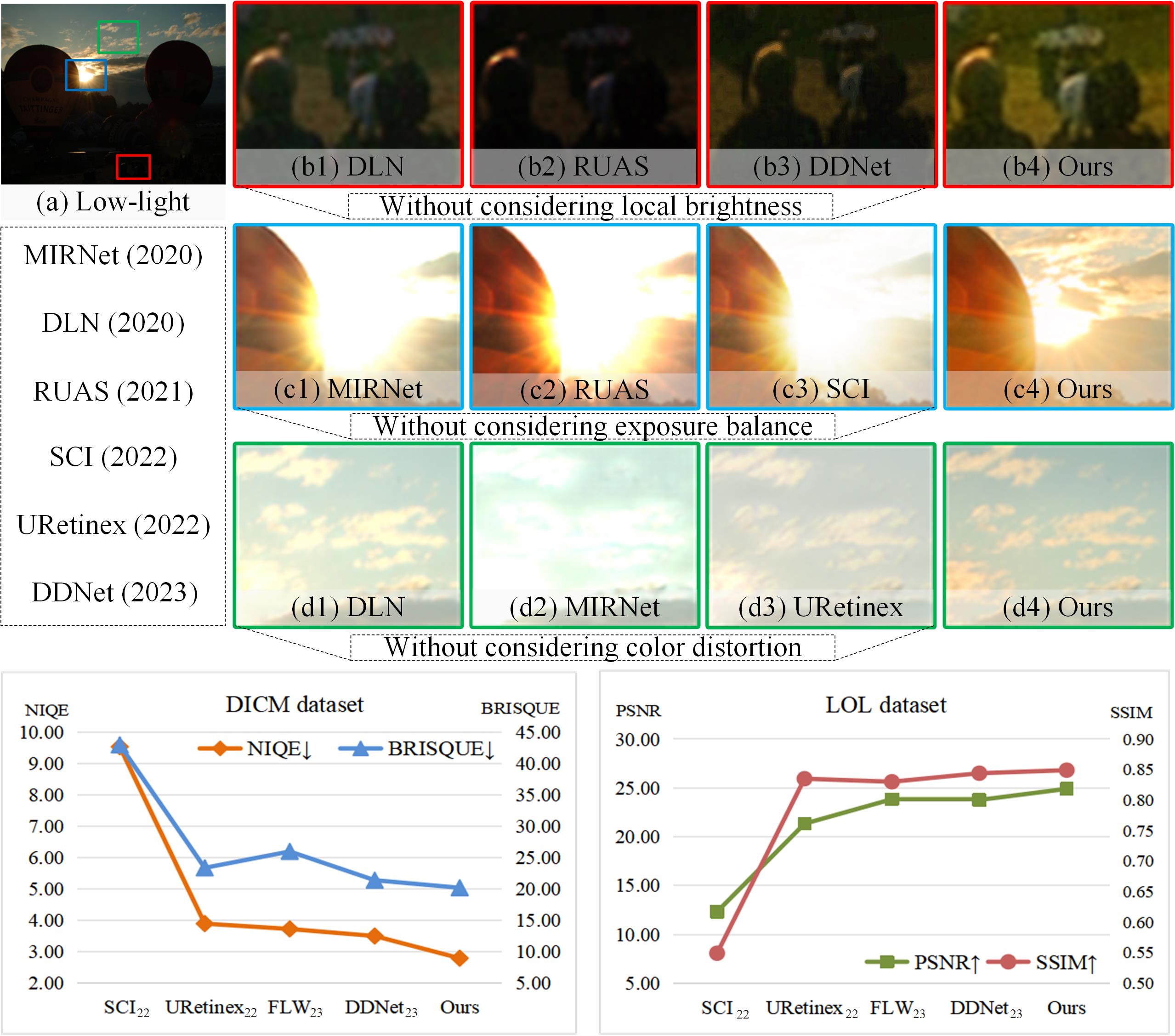}
	\DeclareGraphicsExtensions.
	\caption{Visually display the enhancement results of different methods from the aspects of local brightness, color preservation, and exposure balance. The red box, blue box, and green box represent enlarged regions of the images enhanced by different methods. Our method exhibits subjective visual advantages in terms of brightness, color, and exposure. More importantly, in the quantitative evaluation metrics presented in the line chart, our method achieved the highest scores compared to state-of-the-art methods.}
	\label{fig_int}
\end{figure}

Some representative works in the field of low-light image enhancement have been proposed, which are mainly inspired by histogram equalization \cite{lee2013contrast}, Retinex theory \cite{ma2022low}, and dehazing models \cite{kim2021deep}. However, these methods are designed to focus on brightening the image and increasing the contrast. Unfortunately, they do not pay more attention to the loss of image detail information and global exposure balance of the image, resulting in color distortion, over-enhancement or under-enhancement in the output image.

Data-driven models can learn large-scale priors \cite{zhuang2022underwater, li2023image}, and they have greatly advanced research in low-light image enhancement. Two main categories of existing methods are CNN-based methods \cite{lim2020dslr} and Retinex-based methods \cite{ma2022low}. However, most methods \cite{hao2020low, zamir2020learning, jiang2021enlightengan, wu2022uretinex} extract deep features from the whole image, ignoring the local details and local brightness range of the image, which makes it difficult for the features to be fully utilized. To address this issue, some methods \cite{wang2020lightening, ma2022toward} have also taken into account the illumination balance, but have not considered the comprehensive transformation of details and specific brightness adjustments in the image, the model does not have enough adjustment capacity, resulting in unnatural enhanced images. These methods have insufficient generalization capability, leading to weak adaptation ability in low-light images under different illumination conditions.

\begin{figure*}[!t]
	\centering
	\includegraphics[width=7in]{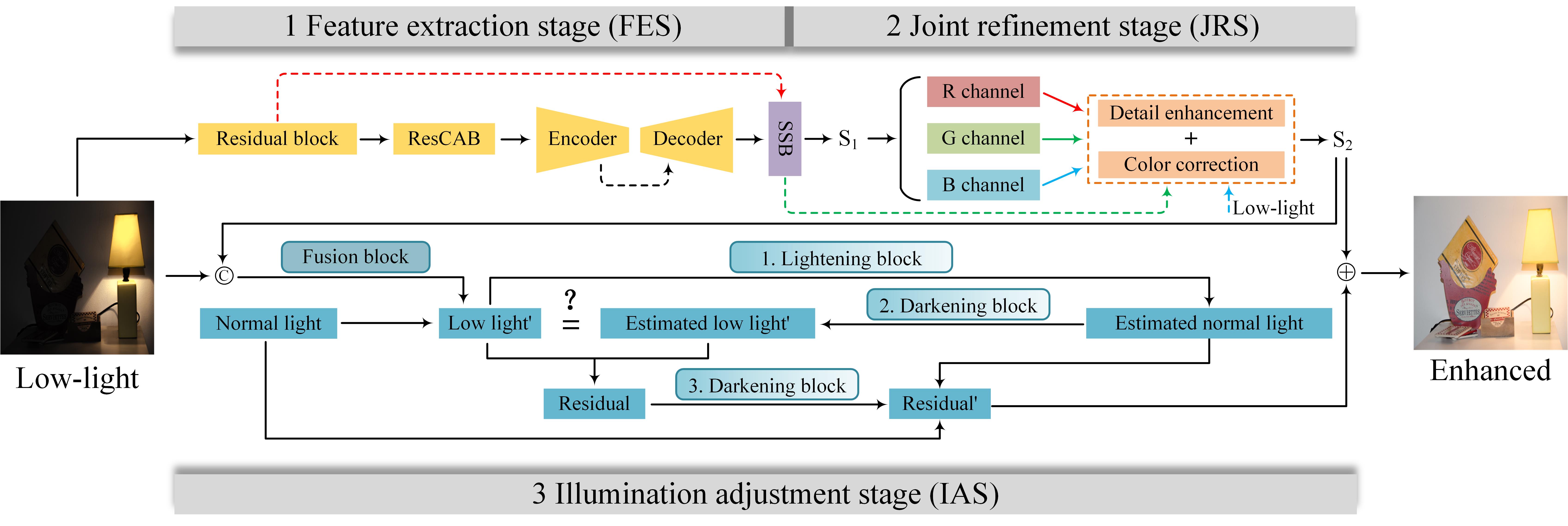}
	\DeclareGraphicsExtensions.
	\caption{Illustration of the proposed Joint Correcting and Refinement Net framework. Firstly, the main features of the input image are extracted in the feature extraction stage (FES). Secondly, detail enhancement especially color correction is performed after the joint refinement stage (JRS). Finally, illumination adjustment stage (IAS) is constructed according to the back projection theory, and the low-light image enhancement is further adjusted using the residuals between the lightening and darkening operations.}
	
	\label{fig_JCRNet1}
\end{figure*}

An example is shown in Fig. \ref{fig_int}. The given low-light image has both underexposed areas and overexposed areas, and exposure balance must be taken into account in the enhancement process. As shown in Fig. \ref{fig_int}(b1), Fig. \ref{fig_int}(b2), and Fig. \ref{fig_int}(b3), the DLN \cite{wang2020lightening}, RUAS \cite{liu2021retinex}, and DDNet \cite{qu2023double}  methods fail to pay attention to the local enhancement of the image, resulting in some regions of the enhanced image still being under-exposed. Additionally, some methods failed to pay attention to the local exposure balance of the image, resulting in over-exposure phenomenon in the originally well-exposed areas, as shown in Fig. \ref{fig_int}(c1), Fig. \ref{fig_int}(c2), and Fig. \ref{fig_int}(c3). If not paying attention to the color of the image, the enhanced image will show distortion effects such as dull color and halo, as shown in Fig. \ref{fig_int}(d1), Fig. \ref{fig_int}(d2), and Fig. \ref{fig_int}(d3). From the above analysis, it can be seen that although existing methods have improved the quality of low-light images in some aspects, they still lack appropriate synergistic mechanisms in terms of brightness, color, and exposure level, making it difficult to maintain a balance in the process of low-light enhancement.

Different from existing methods, firstly, attention and local supervision mechanism can help to extract more comprehensive local information, which is beneficial for improving the model’s learning ability for brightness, color, and illumination. Additionally, cross-stage feature transmission and spatial feature transformation can recover more details and help the fidelity of color information. Finally, inspired by the theory of back-projection, the residual information is gradually learned by using brightening and darkening operations to dynamically adjust the brightness range of images, avoiding over-exposure of enhanced images. Consequently, it is important for low-light image enhancement to have a collaborative correction and refinement mechanism, which can maintain color fidelity and exposure balance while enhancing brightness. Towards this goal, in this paper, we develop a novel approach that maintains the balance of image brightness, color, and illumination. As shown in Fig. \ref{fig_JCRNet1}, the core of our approach is a coordinated three stages network: (a) in the feature extraction stage, residual block cooperative residual channel attention block and encoder-decoder block  for extracting main features, and the self-supervised block (SSB) enables adaptive propagation of useful features, (b) in the joint refinement stage, information is propagated across stages, and image detail enhancement and color distortion correction are realized, (c) in the illumination adjustment stage, according to the back projection theory, the residuals between the normal-light image and the predicted image are learned actively, thereby the exposure balance of the enhanced image is adjusted adaptively. As shown in Fig. \ref{fig_int}(b4), Fig. \ref{fig_int}(c4), and Fig. \ref{fig_int}(d4), the enhanced image of the proposed method with rich color details, balanced exposure, and a pleasant visual experience. Additionally, from the quantitative evaluation line chart provided below in Fig. \ref{fig_int}, it can be observed that our method exhibits comprehensive performance advantages across various metrics.

The main contributions of the proposed method are summarized as follows:

$\bullet$ We propose a novel brightening approach that enables cross-stage feature propagation and adaptive adjustment of image enhancement, while maintaining the coordination of the three stages, to improve the overall image quality.

$\bullet$ We design a joint refinement mechanism for color correction and spatial feature transformation, which plays an important role in recovering the detailed features of the image, especially the color information.

$\bullet$ We introduce a dynamic brightening-darkening block that adjusts the dynamic range of low-light images in a coordinated manner. Simultaneously, the learned residuals adaptively adjust the exposure balance of the enhanced image.

$\bullet$ Extensive experiments on 9 datasets demonstrate that the proposed method has better performance compared to other state-of-the-art methods. Moreover, further experiments have shown that our method has greater advantages in improving the performance of downstream visual tasks such as saliency detection.

\begin{figure*}[!t]
	\centering
	\includegraphics[width=7in]{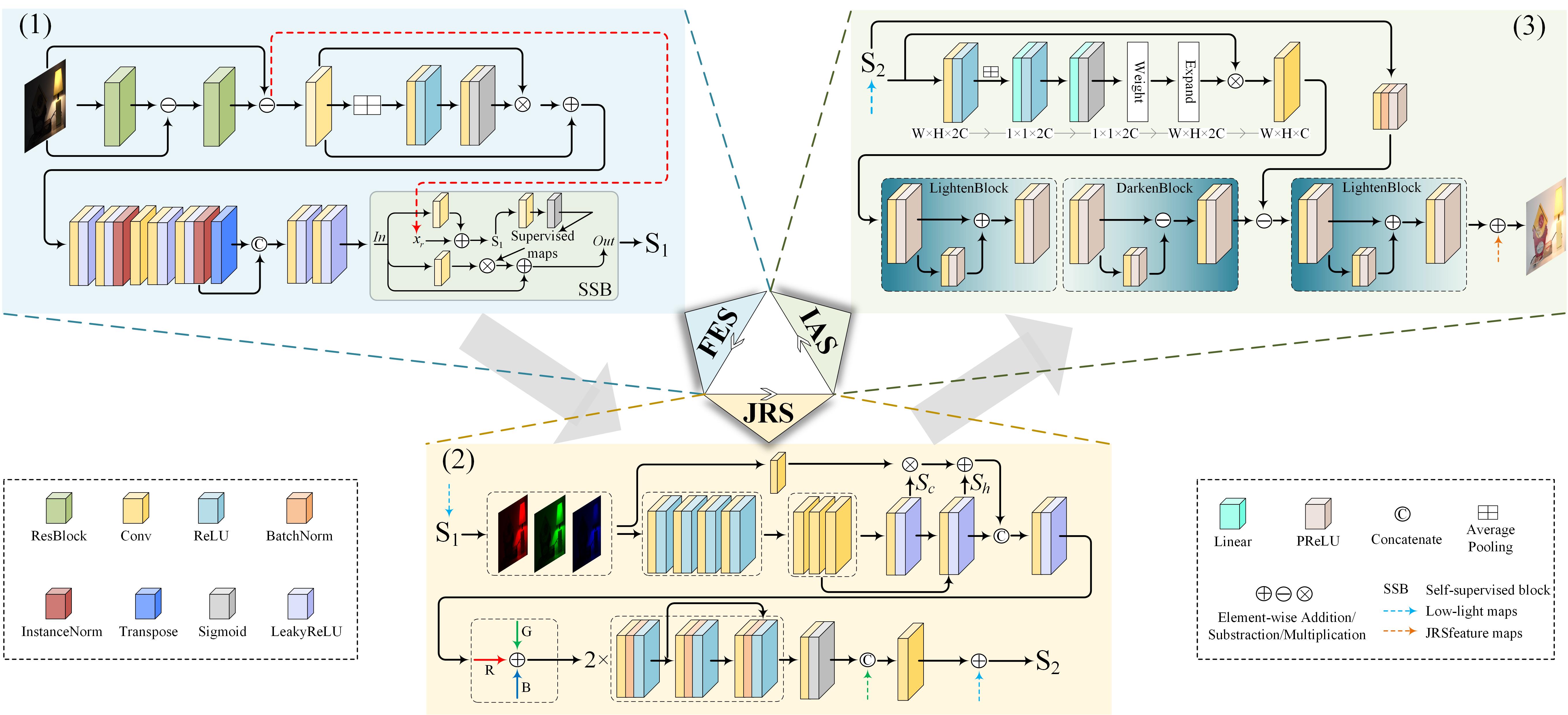}
	\DeclareGraphicsExtensions.
	\caption{Details of the JCRNet architecture. Specifically, the proposed JCRNet consists of three stages: feature extraction stage (FES), joint refinement stage (JRS), and illumination adjustment stage (IAS). The three stages work together to enhance low-light images, thereby balancing brightness, color, and exposure.}
	
	\label{fig_JCRNet2}
\end{figure*}

\section{Related Work}
In the field of computer vision, low-light image enhancement techniques are widely studied and are highly beneficial for high-level tasks. A number of approaches have been developed in recent years to solve this issue, and these have been organized into two broad categories: conventional methods and deep-learning methods.
\subsection{Traditional Methods}
Histogram equalization \cite{yuan2012automatic} is the most commonly used approach. The most representative method is the conventional histogram equalization method proposed by Cheng et al. \cite{cheng2004simple}. However, they under-enhance or over-enhance the image brightness frequently. Therefore, Wang et al. \cite{wang1999image} designed a binary sub-image histogram equalization method for natural exposure processing. On the other hand, to alleviate the problem of detail loss, Lee et al. \cite{lee2013contrast} designed an enhancement method for the hierarchical differential representation of two-dimensional histograms. Nevertheless, these methods introduce a new limitation of color distortion.

Inspired and driven by Retinex theory \cite{land1971lightness}, several enhancement algorithms for low-light images have been developed. Most Retinex-based approaches decompose the low-illumination image into two parts firstly, one for the illumination component and the other for the reflection component. Subsequently, the dynamic range of the estimated illumination is adjusted and recombined with the separated reflection component to generate the final enhancement image. However, to suppress unnecessary noises in the enhancement results, additional denoising techniques should be employed in addition to their decomposition process.

Although traditional methods may slightly enhance the visual quality of low-light images, they mainly focus on increasing contrast, so further refinement is needed to make the colors look more natural.
\subsection{Data-Driven Methods}

In recent years, the ability of deep learning models to learn from data has allowed them to achieve a prominent position in the field of computer vision. It is worth noting that low-light image enhancement works employing collaborative CNNs and Retinex, CNNs, and GANs. Wang et al. \cite{wang2020lightening} proposed a novel deep lightening network (DLN) which views the low-light image enhancement task as a residual learning problem and achieved relatively good results, but did not address the issues of multi-stage feature refinement and image color. Therefore, to conduct multi-stage and multi-scale feature fusion, Zamir et al. \cite{zamir2020learning} developed MIRNet, which can learn richer features in the image. Xu et al. \cite{xu2022snr} proposed the multi-branch SNR-aware transformer to guide feature fusion and also achieved some gains, but they still address issues such as color distortion and uneven illumination.

Subsequently, motivated by the Retinex theory, approaches to estimate the reflection component and illuminance component of low-light images using deep learning models have been proposed successively. Zhang et al. \cite{zhang2021beyond} decomposed the image into two components (KinD++), one responsible for illumination adjustment and the other for degradation removal, and achieved good performance on most of the data. Wu et al. \cite{wu2022uretinex} designed an iterative enhancement model (IAT) based on the Retinex theory, which performed well on real low-light images. Recently, some novel approaches have also been applied to low-light image enhancement tasks. For instance, Wang et al. \cite{wang2023low} ingeniously devised a simple yet highly effective image enhancement scheme based on virtual exposure and image fusion strategies. This method can reveal more information in extremely dark regions and offers flexibility in its application scope. Li et al.\cite{li2023low} designed a gradient-guided knowledge distillation scheme to address the low-light image enhancement problem, achieving a balance between network computational complexity and performance. However, in some challenging data, more attention needs to be paid to issues such as color distortion and dynamic illumination adjustment, otherwise overexposure and other phenomena may occur.

To address the issue of color distortion, Cui et al. \cite{cui2022illumination} proposed a lightweight and fast lighting self-adaptive transformer with color correction. Additionally, Ma et al. \cite{ma2022toward} developed a self-calibrating lighting model that addressed the limitations of data and achieved effective gains. However, these models still did not address the issue of overexposure in some data and had insufficient generalization ability. 
%Overall, the issues addressed by different models are summarized in Table \ref{tab:movti}. Our proposed model addresses both color distortion and overexposure problems in low-light image enhancement.

\section{Proposed Method}
The enhancement framework we construct is shown in Fig. \ref{fig_JCRNet2}. Specifically, the whole network is interwoven by three stages: feature extraction, joint refinement, and illumination adjustment. Among them, the three stages have different divisions of labor and collaborate to jointly enhance low-light images. The proposed method mainly contains several key elements: (a) the residual channel attention block collaborates with the encoder-decoder block to redistribute the image channel weights and extract the main features in low-light images, (b) the joint refinement mechanism to enhance image detail and correct color distortion, which helps to refine visual results, and (c) the lightening-darkening block to learn the residual information between the ground-truth and the predicted image, which can avoid overexposure or underexposure of the enhanced image.

\subsection{Feature Extraction Stage (FES)}

Technically, the main features contained in the shallow layer of the input image are obtained using two residual blocks \cite{mou2022deep} as follows:

\begin{equation}
	{x_r} = x - re{s_2}\left( {re{s_1}(x) - x} \right),
\end{equation}
where, $x$ is original low-light image, $re{s_1}$ and $re{s_2}$ are both a 3×3 convolution and a PReLU activation function. ${x_r}$ is the feature map obtained after two residual blocks.

After two residual blocks processing, the neuron has a larger receptive field and can extract more global information. More importantly, the combination of local fine-grained features and global coarse-grained features results in better low-light enhancement. Therefore, more channel information is extracted when the residual channel attention block reassigns weight to the channels in the feature map. Where, the whole process of attention is divided into three parts: squeeze (${F_{{\rm{sq}}}}$), excitation (${F_{ex}}$), and attention (${F_{scale}}$), the overall process is expressed as follows:

\begin{equation}
	{{\rm{z}}_{\rm{c}}} = {F_{{\rm{sq}}}}({u_c}) = \frac{1}{{H{\rm{ \times }}W}}\sum\limits_{i = 1}^H {\sum\limits_{j = 1}^W {{u_c}(i,j)} },
\end{equation}

\begin{equation}
	s = {F_{ex}}\left( {z,w} \right) = \sigma ({w_2}\delta ({w_{1z}})),
\end{equation}

\begin{equation}
	{\hat x_c} = {F_{scale}}\left( {{u_c},{s_c}} \right) = {u_c} \cdot {s_c}.
\end{equation}

In Equation 2, $H$ and $W$ denote the height and width of the feature map, ${u_c}(i,j)$ denotes the feature information after convolution. In Equation 3, $\delta $ denotes the ReLU activation function, and $\sigma $ denotes the Sigmoid function. $w$, ${w_1}$, and ${w_2}$ all denote weights, where, ${w_1} \in {R^{\frac{c}{r}*C}}$ and ${w_2} \in {R^{\frac{c}{r}*C}} $ respectively. Through the excitation process, the model will learn to obtain the incentive weight, and then assign different weights to each channel, which is the attention process (Equation 4). Finally, different weights are assigned to each channel in the feature map.

%\begin{figure}[!b]
%	\centering
%	\includegraphics[width=0.48\textwidth]{figs/SSB.pdf}
%	\DeclareGraphicsExtensions.
%	\caption{Structure of the self-supervised block (SSB).}
%%	\begin{center}
%%		\flushleft{\textbf{Figure 3: }Structure of the self-supervised block(SSB).}
%%	\end{center}
%	\label{fig_ssb}
%\end{figure}

Especially, to integrate as many residual features as possible, inspired by Resblock, we use the residual channel attention \cite{wang2017residual}. As shown in Fig. \ref{fig_JCRNet2} for FES, there are two skip connections after feature input. One is to make the feature skip output to the later layer directly, allowing low-frequency information to pass through the network faster. The other is to optimize the relationship between channels through channel attention. Finally, the two feature maps are added together, thus achieving residual attention on the channels. Additionally, some deeper information is extracted by the encoder-decoder block. Note that the intermediate information can also be naturally propagated to the decoder, which is beneficial from the skip connection between encoder and decoder. More importantly, at the end of the feature extraction stage, we introduce a self-supervised block (SSB) \cite{waqas2021multi} to generate a supervised feature map between the image features predicted in the first stage and those obtained by residual blocks. Thus, the information useful to the task is extracted in a supervised manner and then passed on to the joint refinement stage. Specifically, the schematic of SSB is illustrated in the bottom right corner of Fig. \ref{fig_JCRNet2}(1).

\subsection{Joint Refinement Stage (JRS)}

After the feature extraction stage, the model can learn the main information features of the original image. However, in some images, the brightness is unevenly distributed and there are areas of extreme darkness. Therefore, details are not well recovered during the enhancement process, resulting in blurred details and color distortion in the final enhanced image. To ameliorate this issue, firstly, the R, G, and B three-channel images are spatially feature transformed for feature-level and spatial-level detail enhancement, and then the color distortion problem \cite{liu2022attention} is corrected according to Retinex theory.

As shown in Fig. \ref{fig_JCRNet2}(2), the feature maps obtained in the feature extraction stage undergo a series of consecutive 3×3 convolutions and ReLU activations separately for each RGB channel to further optimize the features. To maintain the original color tones of the image, the model independently estimates the brightness information for each color channel (R, G, B). The brightness and contrast of each color channel are adjusted using the Retinex decomposition theory to alleviate color distortion issues. The corrected color channels are then merged to reconstruct the post-corrected feature maps.

Connecting the feature maps to the Spatial Feature Transform Layer (SFTL) \cite{wang2018recovering}, further focusing on real details is achieved. In SFTL, inspired by previous research \cite{wang2018recovering}, we use modulation parameters $({S_c}, {S_h})$ to scale and shift each intermediate feature, performing transformations to adjust the response values within the feature maps. This highlights and sharpens subtle features, thereby improving image quality and reducing discrepancies between predicted and ground truth images. Spatial feature transformation introduces learnable parameters, allowing flexibility in associating with enhanced images, enabling the model to learn how to apply spatial transformations effectively, thereby adapting to different types of images. Where, $({S_c}, {S_h}) = R(\Phi)$ and $({S_c}, {S_h} )$ are obtained from prior conditions,

\begin{equation}
	SFTL({S_1}|{S_c}, {S_h} ) = {S_c}  \otimes {S_1} + {S_h},
\end{equation}
where, ${S_1}$ is the image feature maps (R / G / B channel) obtained in the feature extraction stage. ${S_1}$, ${S_c}$, and ${S_h}$ have the same dimension. The modulation parameters are obtained through spatial feature transformation learning, where the learned parameter pairs adaptively influence the output by applying affine transformations in the spatial domain to each intermediate feature map. Specifically, ${S_c}$ represents the learned scaling parameters, and ${S_h}$ represents the learned shifting parameters. $ \otimes $ denotes point-wise multiplication. ${S_c}$ and ${S_h}$ will be described in detail in subsequent ablation experiments.

Based on the Retinex theory \cite{land1971lightness}, we assume that the illumination is estimated independently in each channel, so that the reflectance map can preserve the original hue of the source images. Therefore, we use multiple "convolution + normalization + activation" and two residual blocks to correct the distortion problem. Among them, the size of the convolution kernel in the convolution layer is 3×3, and the number of feature channels of the image in the middle layer is 128. Finally, the image features are changed from 128 channels to 3 channels. Specifically, the image processed in the joint refinement stage can be expressed as:

\begin{equation}
	{x_J} = {x_A}/R(S({x_A},E({x_A})),
\end{equation}
where, the operator / denotes point-wise division, ${x_A}$ denotes the output image processed in the FES. $R$, $S$, and $E$ respectively denote color correction, spatial feature transformation, and detail enhancement.
\subsection{Illumination Adjustment Stage (IAS)}
Inspired by the back projection theory \cite{haris2018deep}, utilizing a darkening operation, a normal-light image can be turned into a low-light image. On the contrary, the aim of low-light image enhancement is to perform a suitable lightening operation to predict the normally illuminated image. Specifically, assuming that the lightening and darkening operations are perfect processes, the predicted image and ground truth are consistent. However, in fact, there is a difference from a normal-light image to a low-light image or from a low-light image to a normal-light image. Their difference denotes the error of a lightening operation or a darkening operation. Therefore, according to the residual information, the error of a lightening operation in enhancing process can be estimated progressively.

The back projection theory plays an important role in solving the exposure imbalance problem by exploiting abundant residual information in the enhancement process. As shown in Fig. \ref{fig_JCRNet2}. Firstly, the original image and the output of the previous stage are employed as the input. Next, these features are aggregated to make more effective use of them, thus improving the expression ability of the features. Specifically, the features in the two input feature maps are concatenated, calibrated, and integrated. Secondly, a lightening block ${L_1}$ is used to predict a normal-light image from a low-light image. Next, a darkening block $D$ predicts a low-light image. Thirdly, the difference between the predicted low-light image and the original low-light image is calculated and denoted as the residual map ${R'_F}$. Similarly, for residual ${R'_F}$, another lightening block ${L_2}$ is used to estimate the residual value ${R_F}$ under normal lighting conditions. Finally, the residual value ${R_F}$ is added to the final prediction map to obtain the enhanced image.

%\begin{figure}[!b]
%	\centering
%	\includegraphics[width=0.45\textwidth]{figs/loss.jpg}
%	\DeclareGraphicsExtensions.
%	\caption{The process of the change in loss from the beginning of training until the network model converges.}
%	%	\begin{center}
%	%		\flushleft{\textbf{Figure 3: }Structure of the self-supervised block(SSB).}
%	%	\end{center}
%	\label{fig_loss}
%\end{figure}

Specifically, the operation of the lightening block includes encoding convolution, offset convolution, and decoding convolution. Firstly, the encoding convolution includes a Convolutional layer and a Parameterized ReLU layer, which reduces the number of feature channels from 64 to 32 to extract more representative features. Subsequently, offset convolution is used to learn the difference between normal-light images and low-light images. Objectively, the normal-light image has a larger pixel value than the low-light image. PReLU is a type of activation function used in deep learning networks to remove the negative values of the offset. Then by using an offset, the image brightness is adjusted by increasing the intensity of each pixel in the image by a specific value. Finally, the number of image feature channels is increased from 32 to 64 using decoded convolution. On the contrary, the darkening block subtracts the offset. The whole process of residual learning is as follows: 

\begin{equation}
	{R_F} = {L_2}(\lambda {x_F} - D({L_1}({x_F}))),
\end{equation}
where $\lambda  \in R$ is weight, used to balance the residual update. ${x_F}$ is the image after the feature aggregation block. ${L_1}$ and ${L_2}$ are lightening blocks, and $D$ is darkening block.

\subsection{Loss Function}
The L1 and L2 loss functions may cause the enhanced image to be too smooth, resulting in the lack of visual realism of the enhanced image, while Charbonnier loss \cite{charbonnier1994two} is more stable. Additionally, the edge loss can well consider the high-frequency texture structure information and improve the image detail performance. The Charbonnier loss and the edge loss are combined as the total loss function of the network model. Where, the Charbonnier loss and the edge loss can be expressed as:

\begin{equation}
	{L_{char}} = \sqrt {{{\left\| {x - {x^{gt}}} \right\|}^2} + {\varepsilon ^2}},
\end{equation}

\begin{equation}
	{L_{edge}} = \sqrt {{{\left\| {\Delta (x) - \Delta {x^{gt}}} \right\|}^2} + {\varepsilon ^2}},
\end{equation}
where, ${L_{char}}$ is Charbonnier loss and ${L_{edge}}$ is edge loss. $x$ is the enhanced image, and ${x^{gt}}$ is the ground-truth. $\varepsilon $ is a constant, usually set to ${10^{ - 3}}$ as a rule of thumb. $\Delta $ is the Laplace operator. In the end, the total loss is:

\begin{equation}
	{L_{total}} = {L_{char}}(x,{x^{gt}}) + \lambda {L_{{\rm{edge}}}}(x,{x^{gt}}),
\end{equation}
where, $\lambda $ is a parameter that controls the weights of ${L_{char}}$ and ${L_{edge}}$, which is set to 0.05 \cite{jiang2020multi} according to experience.

%\begin{figure*}[!t]
%	\centering
%	\includegraphics[width=0.98\textwidth]{figs/fig6.jpg}
%	\DeclareGraphicsExtensions.
%	\begin{center}
%		\flushleft{\textbf{Figure 6: }Visualization of results on the NPE dataset \cite{wang2013naturalness}. The color histogram is used to represent its color distribution.}
%	\end{center}
%	\label{fig_sim}
%\end{figure*}

\section{Experiment}
\subsection{Experimental Settings}

\textbf{Implementation Details.} We implement our models using PyTorch \cite{paszke2017automatic} and perform the experiments on Ubuntu 18.04 operating system, NVIDIA GTX 3090 GPU. Specifically, we employ the Adam optimizer with a cosine annealing strategy \cite{loshchilov2016sgdr}, where the learning rate is decayed from $2 \times {10^{ - 4}}$ to $1 \times {10^{ - 6}}$. We train the network on 256×256 image patches with a batch size of 10. We have also implemented our method using Huawei's open-source framework, MindSpore \cite{Huawei}, while keeping the experimental details unchanged.

\begin{figure*}[!t]
	\centering
	\includegraphics[width=0.95\textwidth]{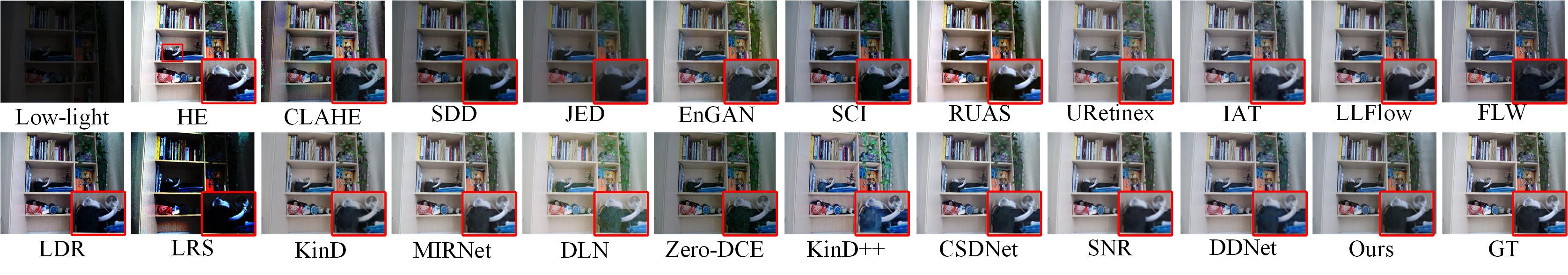}
	\DeclareGraphicsExtensions.
	\caption{Visualizing the results of the LOL dataset \cite{wei2018deep}. For clarity, the magnified parts of the images are displayed.}
	\label{fig_lol}
\end{figure*}

\begin{figure*}[!t]
	\centering
	\includegraphics[width=0.95\textwidth]{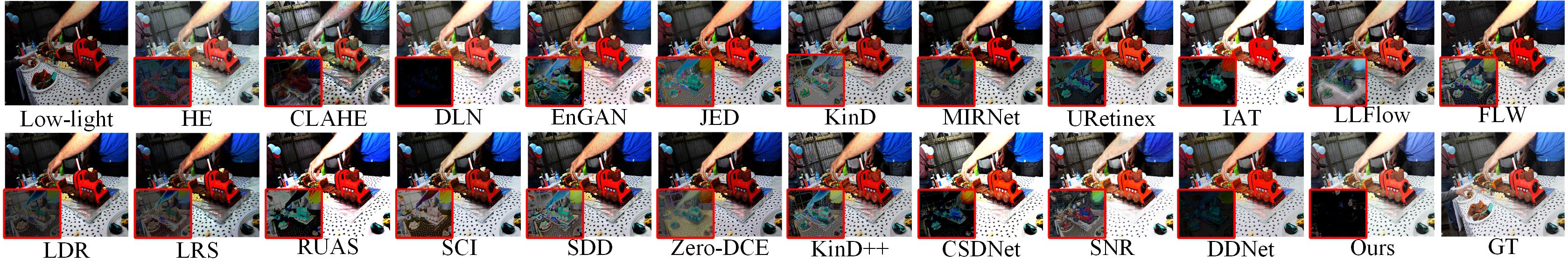}
	\DeclareGraphicsExtensions.
	\caption{Visualizing the results of the COCO dataset \cite{lin2014microsoft}. For a clear comparison of the visual effects, the error images between the ground-truth image and each enhancement result are shown in the lower left corner.}
	\label{fig_coco}
\end{figure*}

\begin{figure*}[!t]
	\centering
	\includegraphics[width=0.95\textwidth]{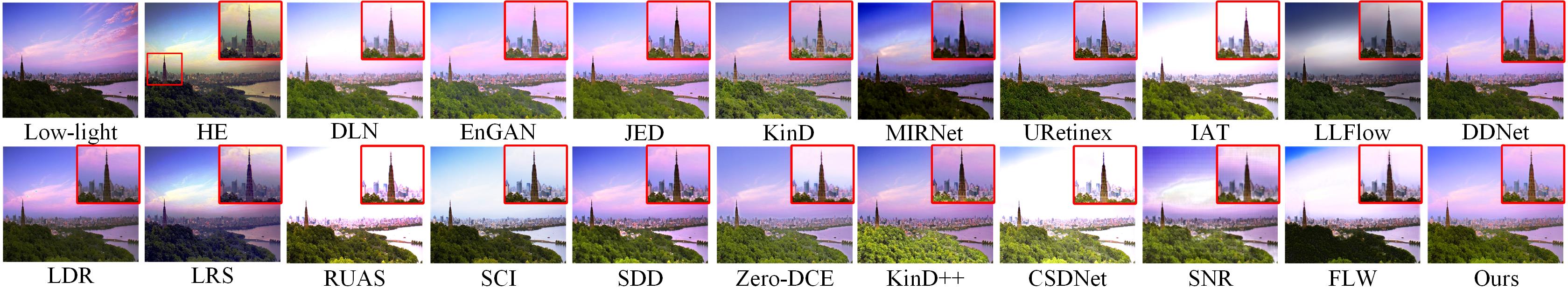}
	\DeclareGraphicsExtensions.
	\caption{Visualizing the results of the NPE dataset \cite{wang2013naturalness}. For clarity, the magnified parts of the images are displayed.}
	\label{fig_npe}
\end{figure*}

\begin{figure*}[!t]
	\centering
	\includegraphics[width=0.95\textwidth]{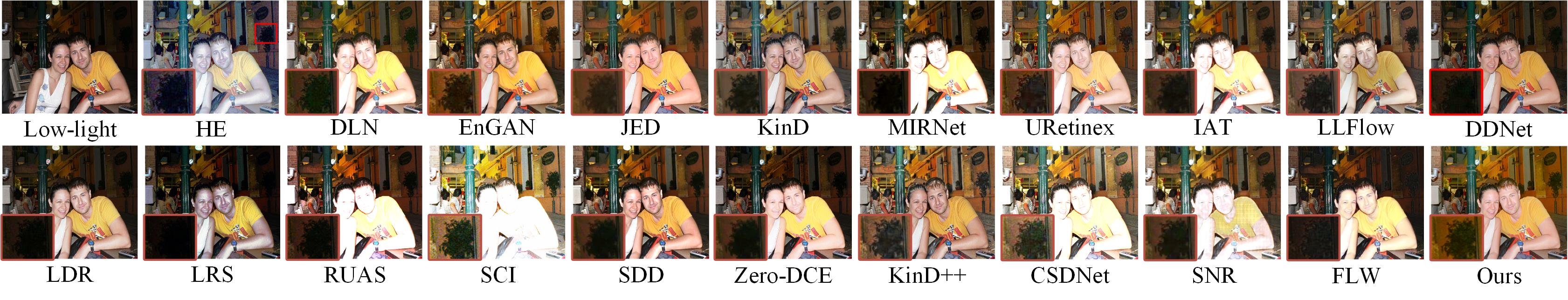}
	\DeclareGraphicsExtensions.
	\caption{Visualizing the results of the VV dataset \cite{vonikakis2018evaluation}. For clarity, the magnified parts of the images are displayed.}
	\label{fig_vv}
\end{figure*}

\begin{figure*}[!t]
	\centering
	\includegraphics[width=0.95\textwidth]{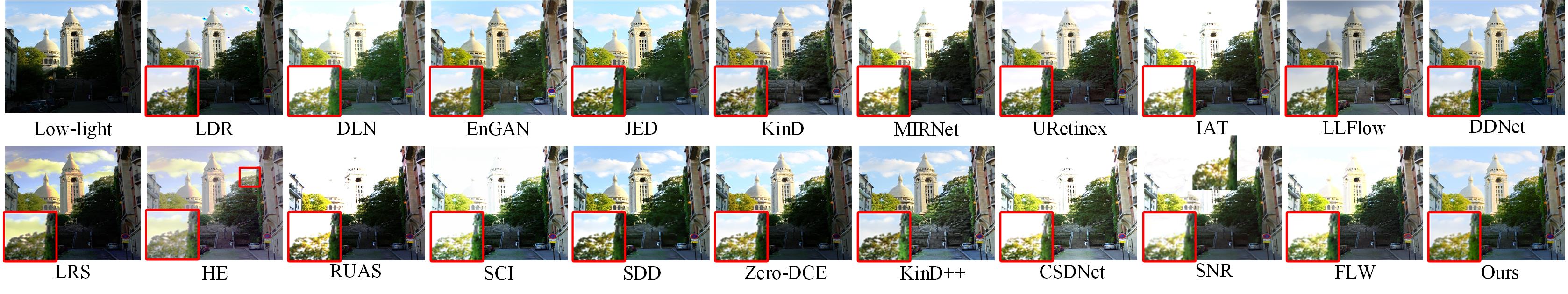}
	\DeclareGraphicsExtensions.
	\caption{Visualizing the results of the Exdark dataset \cite{loh2019getting}. For clarity, the magnified parts of the images are displayed.}
	\label{fig_exdark}
\end{figure*}

\begin{figure*}[!t]
	\centering
	\includegraphics[width=0.95\textwidth]{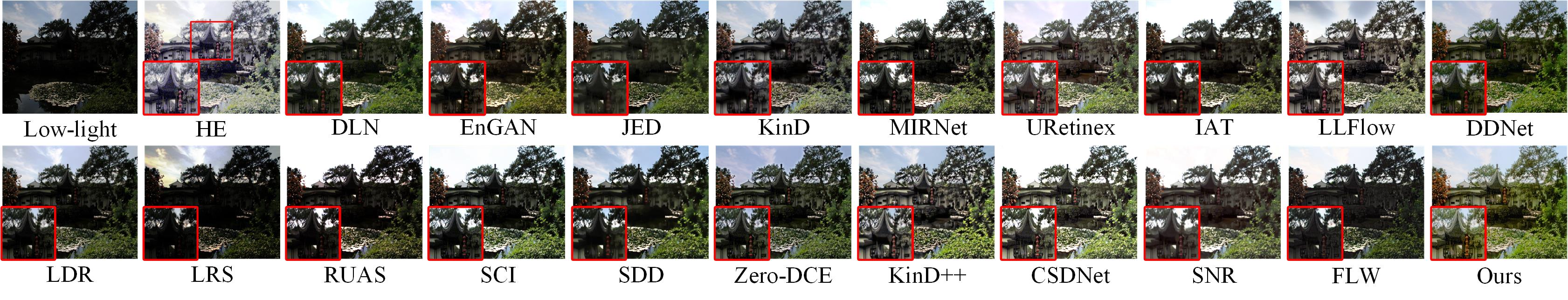}
	\DeclareGraphicsExtensions.
	\caption{Visualizing the results of the MEF dataset \cite{ma2015perceptual}. For clarity, the magnified parts of the images are displayed.}
	\label{fig_mef}
\end{figure*}

\begin{figure*}[!t]
	\centering
	\includegraphics[width=0.95\textwidth]{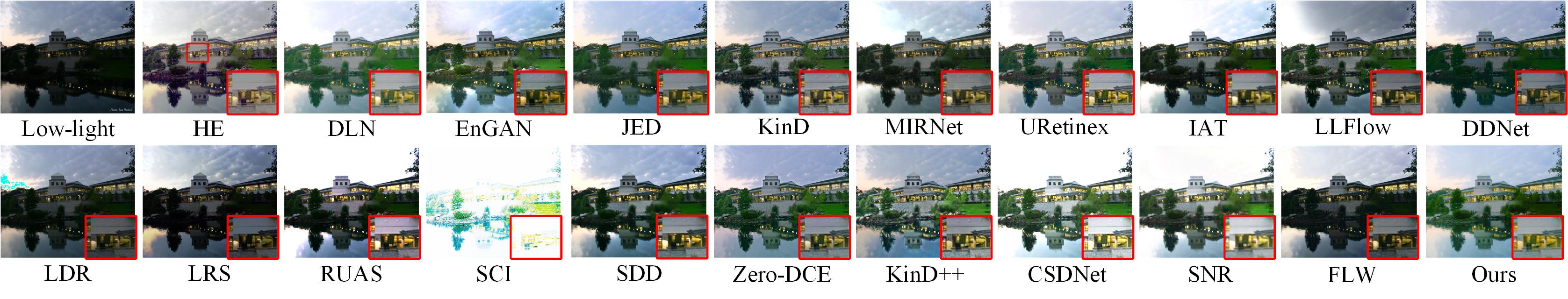}
	\DeclareGraphicsExtensions.
	\caption{Visualizing the results of the DICM dataset \cite{lee2013contrast}. For clarity, the magnified parts of the images are displayed.}
	\label{fig_dicm}
\end{figure*}

\begin{table*}[htbp]
	\centering
		\tabcolsep=0.13cm
	\caption{Average performance comparison between multiple method and the proposed method enhancement on LOL \cite{wei2018deep}, COCO \cite{lin2014microsoft} and MIT \cite{bychkovsky2011learning} datasets. OursPT is based on the PyTorch framework, while OursMS is based on the MindSpore framework. The best, second-best, and third-best scores are represented by red, blue, and green respectively.}
	\scalebox{0.64}{
	\begin{tabular}{c|c|ccccccccccccccccccccc|cc}
		\toprule
		\multicolumn{2}{c|}{\multirow{2}[4]{*}{Methods}} & \multicolumn{4}{c|}{Traditional} & \multicolumn{19}{c}{Deep Learning} \\
		\cmidrule{3-25}    \multicolumn{2}{c|}{} & HE    & CLAHE & LDR   & \multicolumn{1}{c|}{LRS} & SCI   & RUAS  & Zero-DCE & KinD  & MIRNet & EnGAN & DLN   & SDD   & JED   & URetinex & LLFlow & IAT   & SNR   & CSDNet & kinD++ & FLW   & DDNet & OursPT & OursMS \\
		\midrule
		\multirow{5}[2]{*}{LOL} & PSNR↑ & 15.255 & 13.027 & 15.945 & 10.218 & 12.318 & 18.226 & 15.890 & 17.360 & \textcolor[rgb]{ 0,  0,  1}{24.197} & 18.071 & 21.427 & 15.690 & 15.904 & 21.328 & 21.335 & 23.388 & 22.410 & 20.723 & 16.857 & \textcolor[rgb]{ 0,  .69,  .314}{23.840} & 23.760 & \textcolor[rgb]{ 1,  0,  0}{24.901} & 23.563 \\
		& SSIM↑ & 0.485 & 0.466 & 0.547 & 0.418 & 0.549 & 0.717 & 0.550 & 0.680 & 0.832 & 0.688 & 0.764 & 0.627 & 0.663 & 0.835 & \textcolor[rgb]{ 1,  0,  0}{0.883} & 0.814 & 0.824 & 0.837 & 0.752 & 0.830 & 0.844 & \textcolor[rgb]{ 0,  .69,  .314}{0.849} & \textcolor[rgb]{ 0,  0,  1}{0.874} \\
		& BRISQUE↓ & 43.513 & 34.051 & 39.083 & 20.453 & 24.988 & \textcolor[rgb]{ 0,  .69,  .314}{15.595} & 21.780 & 21.420 & 17.066 & 16.538 & 16.245 & 23.795 & 21.852 & 19.667 & 22.094 & 20.244 & 20.300 & 19.609 & 21.367 & 20.501 & 17.475 & \textcolor[rgb]{ 0,  0,  1}{14.001} & \textcolor[rgb]{ 1,  0,  0}{13.946} \\
		& NIQE↓ & 6.506 & 6.636 & 6.382 & 6.545 & 6.617 & 6.867 & 5.170 & 4.971 & 4.657 & 4.813 & \textcolor[rgb]{ 0,  .69,  .314}{4.316} & 5.091 & 5.149 & 4.722 & 5.779 & 4.818 & 5.436 & 4.331 & 4.679 & 5.029 & \textcolor[rgb]{ 1,  0,  0}{4.023} & \textcolor[rgb]{ 0,  0,  1}{4.039} & 4.381 \\
		& LPIPS↓ & 0.264 & 0.309 & 0.139 & 0.451 & 0.212 & 0.354 & 0.718 & 0.463 & 0.099 & 0.647 & 0.177 & 0.206 & 0.205 & 0.135 & 0.108 & 0.104 & 0.095 & 0.102 & 0.220 & 0.120 & \textcolor[rgb]{ 0,  0,  1}{0.084} & \textcolor[rgb]{ 1,  0,  0}{0.082} & \textcolor[rgb]{ 0,  .69,  .314}{0.089} \\   
		\midrule
		\multirow{5}[2]{*}{COCO} & PSNR↑ & 16.395 & 14.507 & 14.232 & 11.497 & 12.613 & 13.884 & 11.937 & 19.257 & 17.291 & 16.583 & \textcolor[rgb]{ 0,  .69,  .314}{23.270} & 14.277 & 18.203 & 20.881 & 16.895 & 17.352 & 20.549 & 17.241 & 18.043 & 20.841 & 19.502 & \textcolor[rgb]{ 1,  0,  0}{24.045} & \textcolor[rgb]{ 0,  0,  1}{23.788} \\
		& SSIM↑ & 0.763 & 0.636 & 0.558 & 0.483 & 0.484 & 0.524 & 0.447 & 0.836 & 0.705 & 0.696 & \textcolor[rgb]{ 0,  0,  1}{0.862} & 0.603 & 0.781 & 0.817 & 0.733 & 0.695 & 0.799 & 0.716 & 0.779 & 0.742 & 0.732 & \textcolor[rgb]{ 1,  0,  0}{0.883} & \textcolor[rgb]{ 0,  .69,  .314}{0.855} \\
		& BRISQUE↓ & 27.273 & 30.408 & 26.404 & 28.535 & 29.904 & 35.784 & 29.569 & 30.037 & 29.214 & 29.140 & 29.214 & 30.348 & \textcolor[rgb]{ 0,  .69,  .314}{22.025} & \textcolor[rgb]{ 0,  0,  1}{21.651} & 25.755 & 24.546 & 27.800 & 27.327 & 25.583 & 24.774 & 26.555 & 24.300 & \textcolor[rgb]{ 1,  0,  0}{20.943} \\
		& NIQE↓ & 4.071 & 4.082 & 4.134 & 4.824 & 8.077 & 4.659 & 4.104 & \textcolor[rgb]{ 0,  0,  1}{3.541} & 4.089 & 3.936 & \textcolor[rgb]{ 0,  .69,  .314}{3.736} & 4.418 & 5.516 & 3.954 & 4.481 & 4.383 & 4.138 & 4.983 & 4.859 & 3.960 & 3.825 & \textcolor[rgb]{ 1,  0,  0}{3.231} & 3.840 \\
		& LPIPS↓ & 0.272 & 0.278 & 0.254 & 0.455 & 0.122 & 0.162 & 0.166 & 0.109 & 0.112 & 0.146 & \textcolor[rgb]{ 0,  0,  1}{0.029} & 0.144 & 0.204 & 0.074 & 0.143 & 0.185 & 0.205 & 0.152 & 0.107 & 0.237 & 0.095 & \textcolor[rgb]{ 1,  0,  0}{0.024} & \textcolor[rgb]{ 0,  .69,  .314}{0.053} \\
		\midrule
		\multirow{5}[2]{*}{MIT} & PSNR↑ & 16.017 & 14.199 & \textcolor[rgb]{ 1,  0,  0}{21.693} & 14.923 & 18.055 & \textcolor[rgb]{ 0,  .69,  .314}{20.830} & 18.690 & 17.090 & 20.051 & 16.297 & 17.698 & 20.370 & 18.370 & 13.888 & 18.011 & 18.318 & 13.075 & 9.241 & 15.043 & 18.768 & 18.308 & \textcolor[rgb]{ 0,  0,  1}{20.920} & 19.806 \\
		& SSIM↑ & 0.758 & 0.633 & 0.834 & 0.641 & \textcolor[rgb]{ 0,  0,  1}{0.861} & 0.854 & 0.830 & 0.770 & 0.768 & 0.797 & 0.817 & 0.810 & 0.750 & 0.720 & 0.792 & 0.785 & 0.694 & 0.666 & 0.761 & 0.797 & 0.776 & \textcolor[rgb]{ 1,  0,  0}{0.866} & \textcolor[rgb]{ 0,  .69,  .314}{0.859} \\
		& BRISQUE↓ & 36.747 & 48.220 & 27.279 & 36.914 & 39.780 & 28.856 & 22.750 & 19.440 & 28.393 & 28.985 & 28.393 & 22.500 & 28.980 & 24.038 & 18.643 & 19.698 & 23.809 & \textcolor[rgb]{ 0,  0,  1}{12.769} & 35.907 & 16.403 & \textcolor[rgb]{ 0,  .69,  .314}{12.946} & \textcolor[rgb]{ 1,  0,  0}{12.398} & 14.778 \\
		& NIQE↓ & \textcolor[rgb]{ 0,  0,  1}{3.667} & 3.743 & 3.669 & 4.040 & 3.963 & 3.812 & 3.858 & 3.839 & 4.064 & \textcolor[rgb]{ 1,  0,  0}{3.624} & \textcolor[rgb]{ 0,  .69,  .314}{3.684} & 4.291 & 4.781 & 4.030 & 4.460 & 4.565 & 4.536 & 4.187 & 3.984 & 4.166 & 3.769 & 3.900 & 4.158 \\
		& LPIPS↓ & 0.166 & 0.254 & 0.273 & 0.367 & 0.171 & \textcolor[rgb]{ 0,  0,  1}{0.141} & 0.276 & 0.229 & 0.166 & 0.209 & 0.162 & 0.628 & \textcolor[rgb]{ 0,  .69,  .314}{0.161} & 0.342 & 0.285 & 0.390 & 0.390 & 0.384 & 0.174 & 0.305 & 0.179 & \textcolor[rgb]{ 1,  0,  0}{0.129} & 0.163 \\
		\bottomrule
	\end{tabular}%
}
	\label{tab:lolcocomit}%
\end{table*}%

\begin{table*}[htbp]
	\centering
	\tabcolsep=0.13cm
	\caption{Average performance comparison between multiple method and the proposed method enhancement on multiple no-reference dataset. OursPT is based on the PyTorch framework, while OursMS is based on the MindSpore framework. The best, second-best, and third-best scores are represented by red, blue, and green respectively. (T: Traditional methods. DL: Deep learning)}
	\scalebox{0.94}{
	\begin{tabular}{c|l|cccccccccccc}
		\toprule
		\multicolumn{2}{c|}{\multirow{2}[4]{*}{Methods}} & \multicolumn{2}{c|}{Exdark dataset} & \multicolumn{2}{c|}{DICM dataset} & \multicolumn{2}{c|}{LIME dataset} & \multicolumn{2}{c|}{MEF dataset} & \multicolumn{2}{c|}{NPE dataset} & \multicolumn{2}{c}{VV dataset} \\
		\cmidrule{3-14}    \multicolumn{2}{c|}{} & BRISQUE↓ & \multicolumn{1}{c|}{NIQE↓} & BRISQUE↓ & \multicolumn{1}{c|}{NIQE↓} & BRISQUE↓ & \multicolumn{1}{c|}{NIQE↓} & BRISQUE↓ & \multicolumn{1}{c|}{NIQE↓} & BRISQUE↓ & \multicolumn{1}{c|}{NIQE↓} & BRISQUE↓ & NIQE↓ \\
		\midrule
		\multirow{4}[2]{*}{T} & HE    & 28.719 & 3.712 & 29.362 & 3.958 & 26.350 & 4.647 & 28.204 & 3.654 & 30.467 & 3.952 & \textcolor[rgb]{ 0,  0,  1}{21.408} & 4.313 \\
		& CLAHE & 39.302 & 3.553 & 21.592 & 3.840 & 27.897 & 4.894 & 20.301 & \textcolor[rgb]{ 0,  0,  1}{3.528} & 32.789 & 4.064 & 21.715 & 4.256 \\
		& LDR   & 32.418 & 3.581 & 36.528 & 3.839 & 26.247 & 4.725 & 26.798 & 3.912 & 28.611 & 4.030 & 26.442 & 4.521 \\
		& LRS   & 45.427 & 4.099 & 21.654 & 4.312 & 21.075 & 5.051 & 21.359 & 4.251 & 28.392 & 4.187 & 30.467 & 4.593 \\
		\midrule
		\multicolumn{1}{c|}{\multirow{16}[4]{*}{DL}} & SCI   & 29.681 & 3.784 & 43.018 & 9.545 & 41.485 & 4.631 & 39.784 & 3.627 & 35.952 & \textcolor[rgb]{ 0,  .69,  .314}{3.579} & 36.932 & 3.870 \\
		& RUAS  & 36.782 & 4.026 & 31.727 & 5.112 & 31.604 & 4.261 & 31.657 & 3.830 & 31.925 & 5.566 & 31.499 & 4.289 \\
		& Zero-DCE & 30.095 & 3.733 & 24.290 & 3.587 & 20.969 & 4.289 & 28.210 & 3.680 & \textcolor[rgb]{ 0,  .69,  .314}{21.839} & 5.602 & 26.830 & 4.869 \\
		& KinD  & 30.240 & 3.567 & 25.802 & 3.510 & 26.718 & 5.344 & 20.218 & 3.876 & 27.090 & 5.519 & 26.247 & 5.043 \\
		& SDD   & 30.004 & 4.057 & 30.027 & 4.606 & 30.026 & 4.939 & 30.135 & 4.314 & 31.491 & 4.992 & 27.511 & 4.234 \\
		& JED   & 30.461 & 4.354 & 30.264 & 3.690 & \textcolor[rgb]{ 0,  0,  1}{19.744} & 5.098 & 30.664 & 4.806 & 31.584 & 4.291 & 26.812 & 4.561 \\
		& URetinex & 29.655 & 3.663 & 23.384 & 3.895 & 22.238 & 4.573 & 19.718 & 3.965 & \textcolor[rgb]{ 1,  0,  0}{19.665} & 4.154 & \textcolor[rgb]{ 0,  .69,  .314}{21.497} & 3.889 \\
		& LLFlow & 29.507 & 4.425 & 21.566 & 4.112 & 26.927 & 4.835 & 29.888 & 4.232 & 43.509 & 4.240 & 27.300 & 3.843 \\
		& IAT   & 29.694 & 4.028 & 26.710 & 4.455 & 29.912 & 4.897 & 26.869 & 4.649 & 43.447 & 4.491 & 22.207 & 4.237 \\
		& SNR   & \textcolor[rgb]{ 1,  0,  0}{26.236} & 3.910 & 23.976 & 4.034 & 26.528 & 5.252 & 26.704 & 4.643 & 25.962 & 4.121 & 24.011 & 3.906 \\
		& CSDNet & 30.118 & 4.496 & 21.737 & 4.804 & 20.617 & 4.920 & 20.088 & 4.107 & 21.997 & 5.709 & 21.846 & 5.021 \\
		& kinD++ & 28.495 & 4.110 & 24.095 & 3.804 & 27.999 & 4.722 & 21.761 & 3.977 & \textcolor[rgb]{ 0,  0,  1}{21.805} & 3.593 & 21.984 & \textcolor[rgb]{ 0,  .69,  .314}{3.306} \\
		& FLW   & 29.556 & 3.859 & 26.003 & 3.724 & 37.103 & 4.434 & 20.991 & 4.198 & 39.189 & \textcolor[rgb]{ 0,  0,  1}{3.567} & 22.090 & 3.510 \\
		& DDNet & 49.883 & \textcolor[rgb]{ 0,  0,  1}{3.391} & \textcolor[rgb]{ 0,  .69,  .314}{21.408} & \textcolor[rgb]{ 0,  .69,  .314}{3.499} & 22.689 & \textcolor[rgb]{ 0,  .69,  .314}{3.969} & \textcolor[rgb]{ 1,  0,  0}{13.604} & 3.801 & 28.986 & 3.668 & 23.446 & 3.542 \\
		\cmidrule{2-14}          & OursPT & \textcolor[rgb]{ 0,  .69,  .314}{28.000} & \textcolor[rgb]{ 1,  0,  0}{3.258} & \textcolor[rgb]{ 0,  0,  1}{20.200} & \textcolor[rgb]{ 1,  0,  0}{2.790} & \textcolor[rgb]{ 0,  .69,  .314}{20.400} & \textcolor[rgb]{ 1,  0,  0}{3.206} & \textcolor[rgb]{ 0,  .69,  .314}{16.500} & \textcolor[rgb]{ 1,  0,  0}{3.220} & 28.300 & \textcolor[rgb]{ 1,  0,  0}{3.315} & \textcolor[rgb]{ 1,  0,  0}{21.100} & \textcolor[rgb]{ 0,  0,  1}{3.061} \\
		& OursMS & \textcolor[rgb]{ 0,  0,  1}{26.721} & \textcolor[rgb]{ 0,  .69,  .314}{3.541} & \textcolor[rgb]{ 1,  0,  0}{19.196} & \textcolor[rgb]{ 0,  0,  1}{3.034} & \textcolor[rgb]{ 1,  0,  0}{17.281} & \textcolor[rgb]{ 0,  0,  1}{3.918} & \textcolor[rgb]{ 0,  0,  1}{14.360} & \textcolor[rgb]{ 0,  .69,  .314}{3.558} & 33.347 & 3.582 & 21.661 & \textcolor[rgb]{ 1,  0,  0}{3.056} \\
		\bottomrule
	\end{tabular}%
}
	\label{tab:multi}%
\end{table*}%

\begin{figure*}[htbp]
	\centering
	\includegraphics[width=0.95\textwidth]{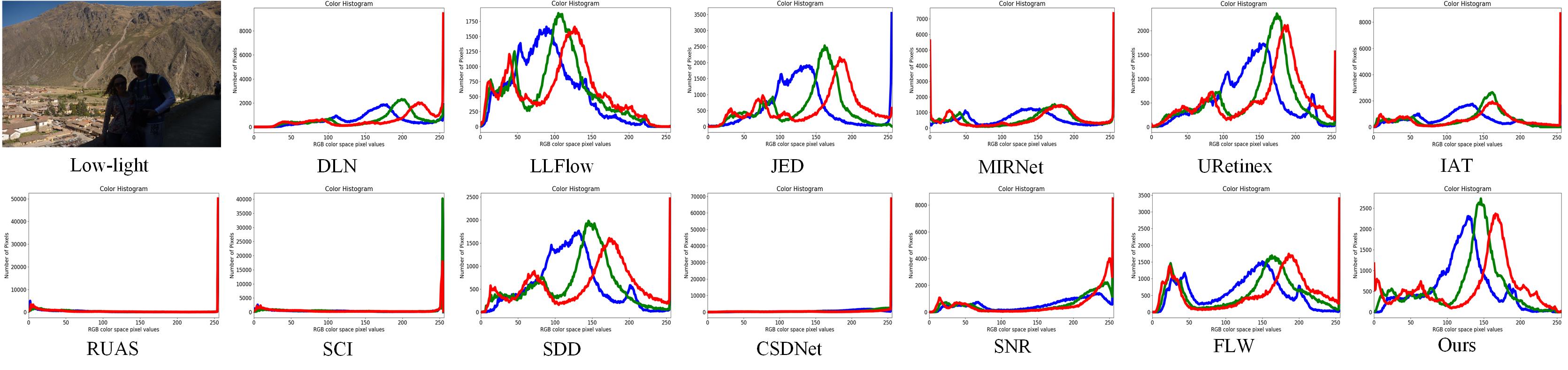}
	\DeclareGraphicsExtensions.
	\caption{Color distribution histograms of enhanced images obtained by different methods. The overall brightness distribution of the enhanced image by the proposed method is better.}
	\label{fig_hist}
\end{figure*}

\begin{figure*}[!t]
	\centering
	\includegraphics[width=1\textwidth]{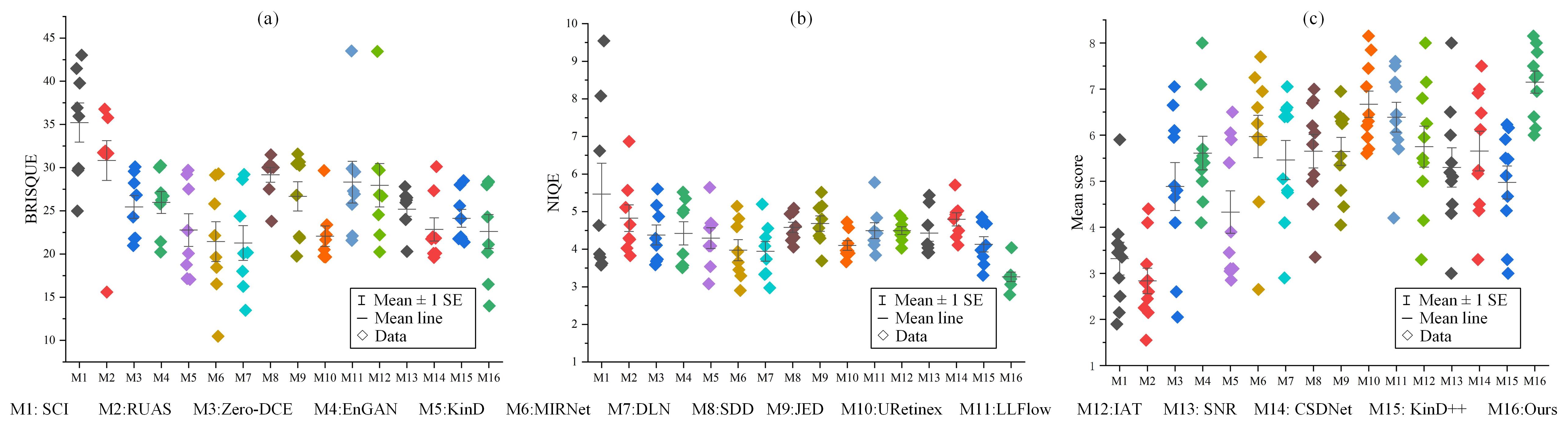}
	\DeclareGraphicsExtensions.
	\caption{(a) shows the mean distribution of BRISQUE for different methods on 8 datasets, (b) shows the mean distribution of NIQE for different methods on 8 datasets, and (c) shows the mean distribution of user study scores obtained from the enhancement results of different methods.}
	\label{fig_us}
\end{figure*}

\textbf{Benchmark Datasets.} Similarly, we train the proposed model using 1000 pairs of images from the Brightening Train dataset \cite{wei2018deep} and 400 pairs of images from the LOL dataset \cite{wei2018deep}. Of these, the remaining 100 pairs of images from the LOL dataset are used for testing. Furthermore, we test the generalization ability of the proposed model on multiple datasets such as COCO (low-light) \cite{lin2014microsoft, sasagawa2020yolo}, MIT \cite{bychkovsky2011learning}, ExDark \cite{loh2019getting}, DICM \cite{lee2013contrast}, LIME \cite{guo2016lime}, MEF \cite{ma2015perceptual}, NPE \cite{wang2013naturalness}, and VV \cite{vonikakis2018evaluation}.

Specifically, the Brightening train dataset consists of 1000 original images from RAISE \cite{dang2015raise}, which are synthesized into low-light images using the interface provided by Adobe Lightroom. The image size is 400×600. The LOL dataset is sourced from Retinex-Net and represents the first dataset of image pairs for low-light enhancement obtained from real-world scenes. The low-light images in the LOL dataset \cite{wei2018deep} are collected by adjusting the ISO through changes in exposure time, encompassing various scenes such as houses, campuses, clubs, and streets. The image size is 400×600. The COCO (low-light) dataset \cite{lin2014microsoft, sasagawa2020yolo} is paired with normal-light images and has been expertly manually annotated. We utilize 959 of these pairs as our test set. The ExDark dataset\cite{loh2019getting} comprises 7363 low-light images captured across a range of conditions, from very low-light environments to twilight (i.e., 10 different conditions). It includes annotations for 12 object classes, akin to PASCAL VOC, at both the image class level and local object bounding boxes. The DICM dataset \cite{lee2013contrast} consists of 69 images collected using commercial digital cameras. The LIME dataset \cite{guo2016lime} encompasses 10 images captured at night. The MEF dataset \cite{ma2015perceptual} comprises 17 images that are the result of multi-exposure fusion in low-light conditions. The NPE dataset \cite{wang2013naturalness} includes 42 low-light images captured in various scenarios, such as cloudy weather, rainy conditions, and twilight. The VV dataset \cite{vonikakis2018evaluation} consists of 26 images taken in backlit or nighttime scenes. These images vary in size among the different test sets, and for our experiments, we uniformly resized them to 384×384.

\textbf{Evaluation Metrics.} We use Peak Signal-to-Noise Ratio (PSNR), Structural Similarity Index Measurement (SSIM) \cite{wang2004image}, and Learned Perceptual Image Patch Similarity (LPIPS) \cite{zhang2018unreasonable} to evaluate the performance of image enhancement. However, for the unpaired data, we evaluate using two non-reference evaluation metrics (BRISQUE \cite{mittal2012no}, NIQE \cite{mittal2012making}).

\subsection{Visual Image Analysis}
The enhanced results of the proposed model are compared with multiple popular approaches, which include traditional histogram equalization-based methods such as HE \cite{krutsch2011histogram}, CLAHE \cite{park2008contrast}, LDR \cite{lee2013contrast}, LRS \cite{srinivasan2006adaptive}, and deep learning-based methods like JED \cite{ren2018joint}, SDD \cite{hao2020low}, DLN \cite{wang2020lightening}, EnGAN \cite{jiang2021enlightengan}, MIRNet \cite{zamir2020learning}, KinD \cite{zhang2019kindling}, Zero-DCE \cite{guo2020zero}, SCI \cite{ma2022toward}, RUAS \cite{liu2021retinex}, URetinex \cite{wu2022uretinex}, LLFlow \cite{wang2022low}, IAT \cite{cui2022illumination}, SNR \cite{xu2022snr}, CSDNet  \cite{ma2021learning}, KinD++  \cite{zhang2021beyond}, FLW \cite{zhang2023fast} and DDNet \cite{qu2023double}. Visual comparison results on the LOL and COCO datasets are shown in Fig. \ref{fig_lol} and Fig. \ref{fig_coco}. It can be seen from Fig. \ref{fig_lol} that the images enhanced by LLFlow and our model are visually better than others. The enhanced image has a relatively natural overall brightness and color. However, the enhanced results of other methods have the problem of overall darkening or overexposure, and due to the excessive brightening of the image or the poor generalization ability of the model, noise appears in some areas. Additionally, we zoom in some details below the enhanced results of each method, which shows that our method is still competitive in terms of details. As shown in Fig. \ref{fig_coco}, in the low-light data of the COCO dataset, the saturation of the proposed method may not be very high in the red region. However, in comparison, in terms of color and detail, the enhanced images of the proposed model perform better. To more clearly show the difference in the enhancement effect of each method, we generate the residual image between each enhancement result and ground truth, which is displayed in the lower left corner. It can be seen in the visual comparison experiments that the enhancement results of DLN and JCRNet have less deviation from the ground truth.

\begin{figure*}[!t]
	\centering
	\includegraphics[width=0.95\textwidth]{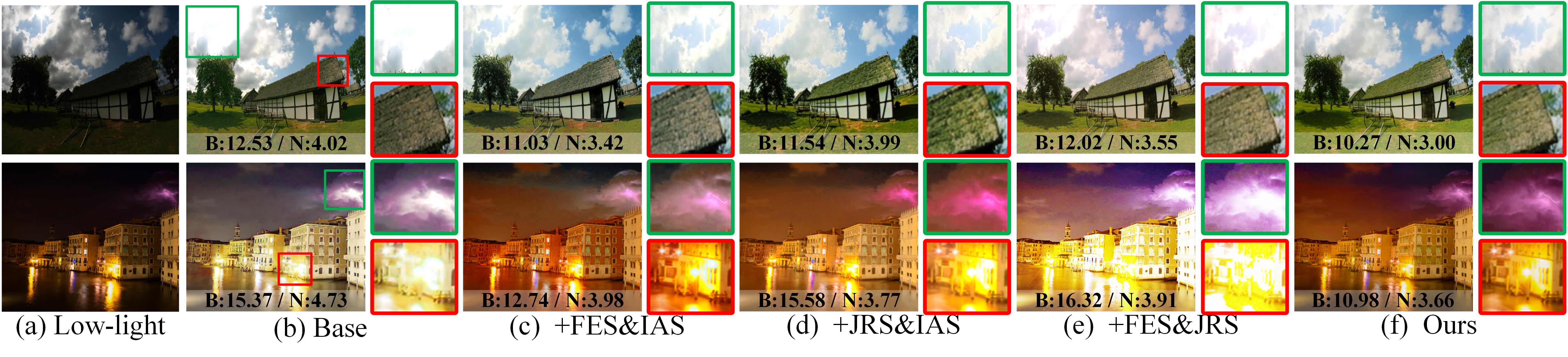}
	\DeclareGraphicsExtensions.
	\caption{Comparison of models with different stages in terms of their subjective visual effects. 'Base' consists of a basic attention and an encoder-decoder. N: NIQE, B:BRISQUE. Smaller values of these metrics indicate that the enhanced images are more natural.}
	\label{fig_xr_stage}
\end{figure*}

\begin{figure*}[!t]
	\centering
	\includegraphics[width=0.95\textwidth]{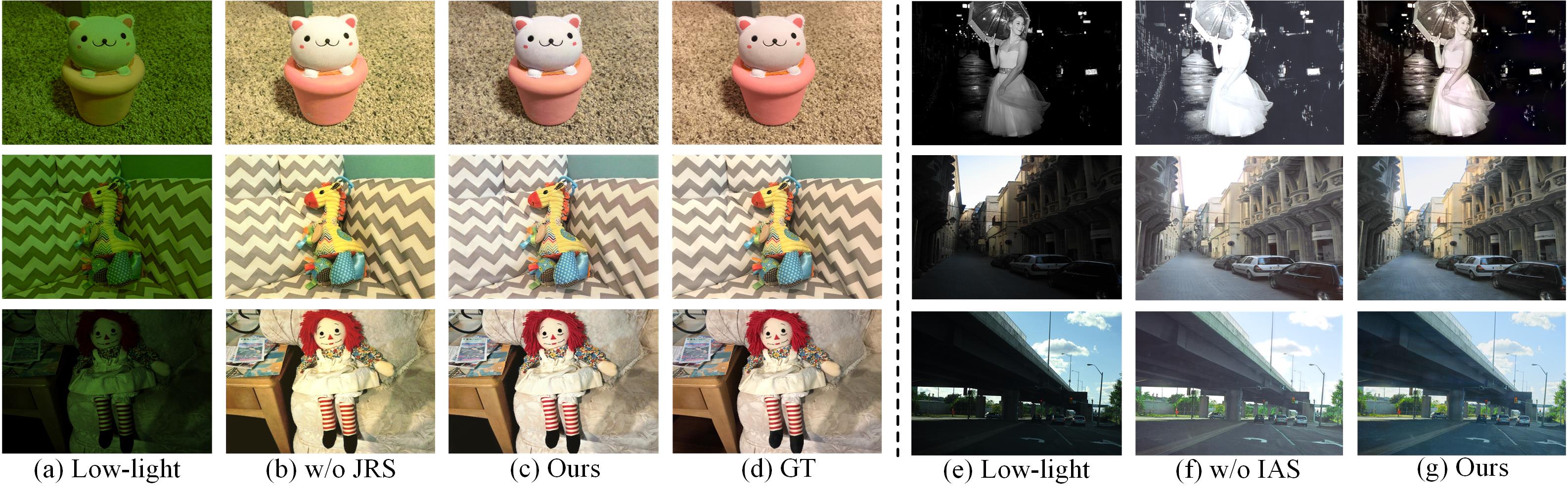}
	\DeclareGraphicsExtensions.
	\caption{Visual comparison of ablation studies on JRS and IAS.}
	%	\begin{center}
	%		\flushleft{\textbf{Figure 10: }Visual comparison of ablation studies on JRS.}
	%	\end{center}
	\label{fig_xr_jrs}
\end{figure*}

\begin{figure*}[!t]
	\centering
	\includegraphics[width=0.95\textwidth]{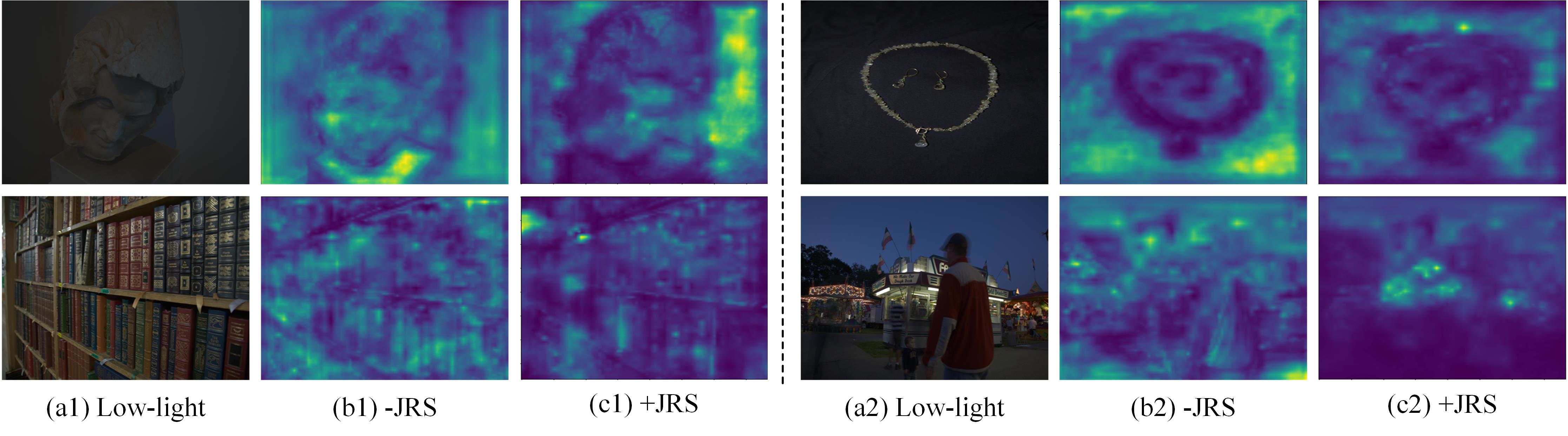}
	\DeclareGraphicsExtensions.
	\caption{Visualize the efficiency of the JRS module. (a1) and (a2) are original low-light images, (b1) and (b2) are spatial distance transformation heatmaps between the enhanced images without the joint refinement module and the ground truth images, (c1) and (c2) are spatial distance transformation heatmaps between the enhanced images with the joint refinement module and the ground truth images. Each pixel in the heatmap contains the perceptual difference value between the two images at the corresponding position. Brighter regions in the heatmaps indicate greater differences.}
	\label{dis-hot}
\end{figure*}

\begin{figure*}[!t]
	\centering
	\includegraphics[width=0.94\textwidth]{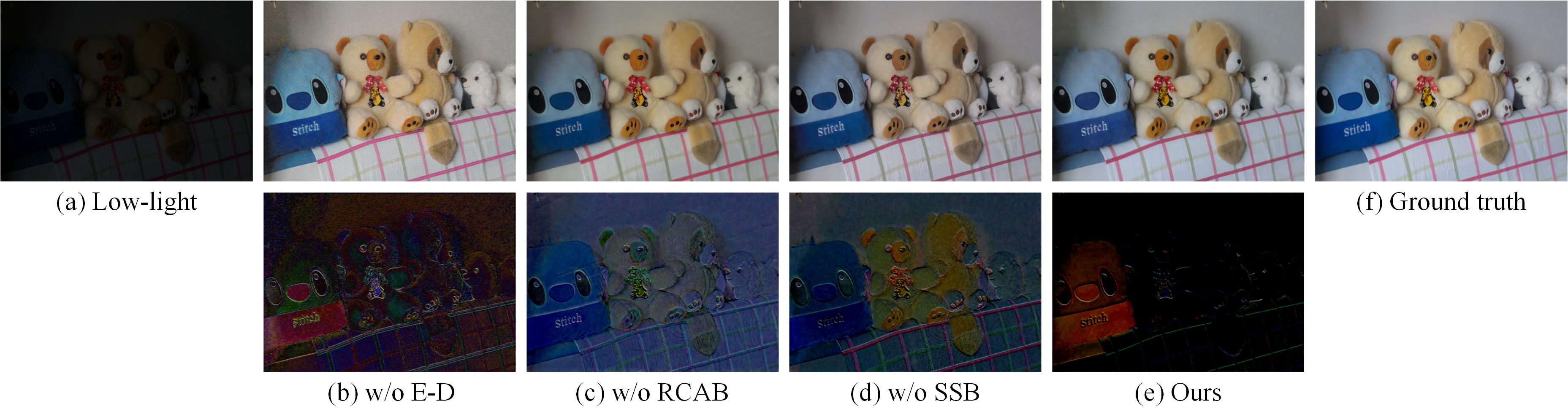}
	\DeclareGraphicsExtensions.
	\caption{Visual comparison of the enhancement results of different components in FES. We show the error images of each image with ground truth for a clear comparison of the visual effects, which can be zoomed to view the details. In order to make the difference between them more visually prominent, we adjust the brightness of the residual map to 150 and the contrast to -50. Our method has fewer residuals compared to the ground truth.}
	\label{fig_xr_componets}
\end{figure*}

Additionally, most datasets have no ground truth paired with low-light images. Fig. \ref{fig_npe} - Fig. \ref{fig_dicm} show the visual comparison results on the NPE, VV, ExDark, MEF, and DICM datasets. Fig. \ref{fig_npe} shows the enhanced image results in the NPE dataset. The enhanced image with JCRNet has a moderate brightness on the whole and is clearer in details (enlarged areas). Fig. \ref{fig_vv} shows the enhancement results for the indoor low-light scenes in the VV dataset. As a whole, most of the enhancement results are overexposed, but only the KinD, EnGAN, DDNet and the proposed methods do not bring about overexposure. Additionally, the JCRNet is superior in terms of color in the enlarged area. Fig. \ref{fig_exdark} shows an example of the results of the experiments performed on the ExDark dataset. The image enhancement results and enlarged details are shown. Fig. \ref{fig_mef} presents examples of low-light image enhancement from the MEF dataset. It is evident from the comparison that our method produces overall brighter results and recovers more color and detail information. Fig. \ref{fig_dicm} visually showcases enhanced images from the DICM dataset. The proposed method and the EnGAN method achieve more satisfactory visual results. Other methods exhibit issues such as overexposure in the sky region, overall dull colors, and unclear details, as seen in SCI, SNR, and IAT, among others. Overall, the proposed method is more competitive. More importantly, the second low-light image is very challenging. In magnified detail areas, most methods have poor enhancement effects. In addition, Fig. \ref{fig_hist} also displays the color histogram distributions of enhanced images obtained by some comparative methods. The color distribution histograms of the enhanced images using the proposed method are relatively better. Therefore, in many aspects, the enhancement results of the proposed method are relatively better.

\subsection{Quantitative Evaluations}

Low-light images from the LOL, COCO, and MIT datasets all have corresponding ground truth pairs. We assess the performance of various algorithms and the naturalness of enhanced images using both reference evaluation metrics (PSNR, SSIM, and LPIPS) and no-reference evaluation metrics (BRISQUE and NIQE). Firstly, Table \ref{tab:lolcocomit} presents quantitative comparisons between the proposed method and state-of-the-art methods of various types. Specifically, Table \ref{tab:lolcocomit} displays the performance evaluation results of various methods, including the JCRNet method, on the LOL test set. The best, second-best, and third-best scores are represented by red, blue, and green respectively. In terms of NIQE, the proposed method slightly outperforms comparison methods such as KinD, MIRNet, EnGAN, DLN, and URetinex. Compared to the existing best-performing methods, the JCRNet method improves PSNR by 0.70 dB. For the Structural Similarity Index (SSIM) metric, the proposed method achieves a score of 0.849, second only to the LLFlow method's optimal score of 0.883, indicating excellent performance in brightness, contrast, and structural aspects of the enhanced images. Additionally, the proposed method employs an iterative approach with dynamic adjustment through back-projection to avoid severe overexposure issues, aligning better with human visual characteristics. Therefore, it also achieves highly competitive scores on three perceptual and naturalness metrics (BRISQUE, NIQE, and LPIPS).

Secondly, in Table \ref{tab:lolcocomit}, we report quantitative evaluation scores for the COCO dataset (low-light data). In COCO dataset, the proposed method outperforms most deep learning-based methods in most metrics, with URetinex and JED methods performing best on the BRISQUE metric. Furthermore, compared to previous state-of-the-art methods, our method improves PSNR by 0.78 dB and increases SSIM by 1.8\%. The metric scores indicate that the proposed method performs well in terms of peak signal-to-noise ratio, image brightness, contrast, and structure. Due to the complexity of image scenes in the COCO dataset and the fact that corresponding normal-light images are manually annotated, resulting in relatively high saturation, our method may not be optimal in several aspects based on naturalness metrics, such as BRISQUE, where it does not perform well. However, our method reduces the LPIPS score by 17\%, indicating that it aligns to some extent with human perception.

Lastly, in Table \ref{tab:lolcocomit}, we report quantitative evaluation scores for the MIT dataset. It is worth noting that the label data in the MIT dataset has been meticulously curated by experts, resulting in high saturation and vivid colors in the ground-truth images. Additionally, there are significant differences in data scenes and distributions between the MIT dataset and the training dataset. Therefore, as evident from the evaluation scores provided in the table, most methods perform relatively poorly in terms of PSNR and SSIM values, especially some traditional methods. However, despite these challenges, the LDR method, which uniformly brightens the entire image, holds an advantage when dealing with low-light data of the MIT type, achieving the highest PSNR score. Furthermore, the proposed method outperforms most other methods in terms of the structural similarity index (SSIM) metric.

For datasets that lack corresponding normal-light images, we employed two widely used no-reference image quality assessment metrics. In Table \ref{tab:multi}, we report the evaluation results of different methods on the ExDark, DICM, LIME, MEF, NPE, and VV datasets. From the reported results, it can be observed that on the ExDark dataset, our proposed JCRNet and the latest DDNet method achieved NIQE scores of 3.258 and 3.391, respectively, indicating that the enhanced images are overall more natural. On the DICM, MEF, and VV datasets, our method exhibits significant improvements in NIQE scores compared to the second-best-performing methods, suggesting that our proposed approach delivers more consistent enhancement performance on these datasets. While on the LIME dataset, although our method does not have a substantial gap in BRISQUE scores, its advantage is more pronounced in NIQE scores. In general, JCRNet achieves a more natural effect at the statistical level. Importantly, compared to other methods, JCRNet can adapt to a wider range of low-light image types and performs best on most datasets.

% Table generated by Excel2LaTeX from sheet 'SCRNet'
\begin{table*}[htbp]
	\centering
	\caption{Ablation study (using the method of controlled variables) on different datasets for different stages of the proposed JCRNet. Best and second best scores are red and blue. (N: NIQE and B: BRISQUE)}
    \scalebox{1}{
	\begin{tabular}{c|ccccccccccccccc}
		\toprule
		\multirow{2}[4]{*}{Model} & \multirow{2}[4]{*}{FES} & \multirow{2}[4]{*}{JRS} & \multicolumn{1}{c|}{\multirow{2}[4]{*}{IAS}} & \multicolumn{2}{c|}{Exdark dataset} & \multicolumn{2}{c|}{DICM dataset} & \multicolumn{2}{c|}{LIME dataset} & \multicolumn{2}{c|}{MEF dataset} & \multicolumn{2}{c|}{NPE dataset} & \multicolumn{2}{c}{VV dataset} \\
		\cmidrule{5-16}          &       &       & \multicolumn{1}{c|}{} & B↓ & \multicolumn{1}{c|}{N↓} & B↓ & \multicolumn{1}{c|}{N↓} & B↓ & \multicolumn{1}{c|}{N↓} & B↓ & \multicolumn{1}{c|}{N↓} & B↓ & \multicolumn{1}{c|}{N↓} & B↓ & N↓ \\
		\midrule
		w/o FES &       & \checkmark     & \checkmark     & \textcolor[rgb]{ 1,  0,  0}{27.638} & \textcolor[rgb]{ 0,  0,  1}{3.474 } & \textcolor[rgb]{ 1,  0,  0}{20.827} & 3.721  & 30.937  & \textcolor[rgb]{ 0,  0,  1}{4.003} & 21.871  & 4.133  & 29.800  & \textcolor[rgb]{ 0,  0,  1}{3.630} & 23.398  & \textcolor[rgb]{ 0,  0,  1}{3.455}  \\
		w/o JRS & \checkmark     &       & \checkmark     & 29.311  & 4.045  & 23.744  & \textcolor[rgb]{ 0,  0,  1}{3.535} & \textcolor[rgb]{ 0,  0,  1}{30.093} & 4.415  & \textcolor[rgb]{ 0,  0,  1}{20.108} & \textcolor[rgb]{ 0,  0,  1}{4.036} & 30.520  & 3.763  & \textcolor[rgb]{ 0,  0,  1}{23.019} & 3.725  \\
		w/o IAS & \checkmark     & \checkmark     &       & 31.653  & 3.529  & 21.811  & 3.595  & 31.503  & 4.303  & 23.461  & 4.329  & \textcolor[rgb]{ 0,  0,  1}{29.403} & 3.742  & 24.237  & 3.740  \\
		Ours  & \checkmark     & \checkmark     & \checkmark     & \textcolor[rgb]{ 0,  0,  1}{28.000} & \textcolor[rgb]{ 1,  0,  0}{3.258} & \textcolor[rgb]{ 0,  0,  1}{20.200} & \textcolor[rgb]{ 1,  0,  0}{2.790} & \textcolor[rgb]{ 1,  0,  0}{28.400} & \textcolor[rgb]{ 1,  0,  0}{3.206} & \textcolor[rgb]{ 1,  0,  0}{16.500} & \textcolor[rgb]{ 1,  0,  0}{3.220} & \textcolor[rgb]{ 1,  0,  0}{28.300} & \textcolor[rgb]{ 1,  0,  0}{3.315} & \textcolor[rgb]{ 1,  0,  0}{21.100} & \textcolor[rgb]{ 1,  0,  0}{3.061} \\
		\bottomrule
	\end{tabular}%
}
	\label{tab:xr-noref}%
\end{table*}%

% Table generated by Excel2LaTeX from sheet 'SCRNet'
\begin{table*}[!t]
	\centering
	\tabcolsep=0.13cm
	\caption{Ablation study (using the method of controlled variables) on LOL and COCO datasets for different stages of the proposed JCRNet. Best and second best scores (PSNR, SSIM, NIQE and BRISQUE) are red and blue.}
	\scalebox{1.35}{
	\begin{tabular}{c|ccccccccccc}
		\toprule
		\multirow{2}[4]{*}{Model} & \multirow{2}[4]{*}{FES} & \multirow{2}[4]{*}{JRS} & \multicolumn{1}{c|}{\multirow{2}[4]{*}{IAS}} & \multicolumn{4}{c|}{LOL dataset} & \multicolumn{4}{c}{COCO dataset} \\
		\cmidrule{5-12}          &       &       & \multicolumn{1}{c|}{} & PSNR↑ & SSIM↑ & BRISQUE↓ & \multicolumn{1}{c|}{NIQE↓} & PSNR↑ & SSIM↑ & BRISQUE↓ & NIQE↓ \\
		\midrule
		w/o FES &       & \checkmark     & \checkmark     & 23.113  & 0.820  & 17.598  & \textcolor[rgb]{ 0,  0,  1}{4.516} & \textcolor[rgb]{ 0,  0,  1}{23.419} & 0.831  & \textcolor[rgb]{ 0,  0,  1}{24.993} & \textcolor[rgb]{ 0,  0,  1}{3.562}  \\
		w/o JRS & \checkmark     &       & \checkmark     & 22.926 & 0.815 & \textcolor[rgb]{ 0,  0,  1}{16.980} & 4.703 & 22.649 & 0.828 & 27.766 & 3.951  \\
		w/o IAS & \checkmark     & \checkmark     &       & \textcolor[rgb]{ 0,  0,  1}{23.318} & \textcolor[rgb]{ 0,  0,  1}{0.833} & 17.610 & 4.586 & 23.318 & \textcolor[rgb]{ 0,  0,  1}{0.859} & 26.343 & 3.669  \\
		Ours  & \checkmark     & \checkmark     & \checkmark     & \textcolor[rgb]{ 1,  0,  0}{24.901} & \textcolor[rgb]{ 1,  0,  0}{0.849} & \textcolor[rgb]{ 1,  0,  0}{14.001} & \textcolor[rgb]{ 1,  0,  0}{4.039} & \textcolor[rgb]{ 1,  0,  0}{24.045} & \textcolor[rgb]{ 1,  0,  0}{0.883} & \textcolor[rgb]{ 1,  0,  0}{24.300} & \textcolor[rgb]{ 1,  0,  0}{3.231} \\
		\bottomrule
	\end{tabular}%
}
	\label{tab:xr-ref}%
\end{table*}%

% Table generated by Excel2LaTeX from sheet 'SCRNet'
\begin{table}[htbp]
	\centering
	\caption{Ablation study on JRS and different components from FES of the proposed JCRNet. Best scores are red.}
	\scalebox{0.97}{
	\begin{tabular}{c|lcccc}
		\toprule
		Dataset & Model & NIQE↓ & BRISQUE↓ & PSNR↑ & SSIM↑ \\
		\multirow{2}[1]{*}{Toys} & w/o JRS & 5.027 & 20.617 & 23.847 & 0.826 \\
		& Full  & \textcolor[rgb]{ 1,  0,  0}{4.642} & \textcolor[rgb]{ 1,  0,  0}{17.959} & \textcolor[rgb]{ 1,  0,  0}{24.524} & \textcolor[rgb]{ 1,  0,  0}{0.848} \\
		\midrule
		\multirow{5}[1]{*}{LOL} & w/o E-D & 4.689 & 14.347 & 24.647 & 0.820 \\
		& w/o CAB & 4.719 & 14.283 & 24.758 & 0.831 \\
		& w/o SSB & 4.514 & 14.572 & 24.531 & 0.830 \\
		& Full  & \textcolor[rgb]{ 1,  0,  0}{4.039} & \textcolor[rgb]{ 1,  0,  0}{14.001} & \textcolor[rgb]{ 1,  0,  0}{24.901} & \textcolor[rgb]{ 1,  0,  0}{0.849} \\
		\midrule
	\end{tabular}%
}
	\label{tab:xr-fes}%
\end{table}%

% Table generated by Excel2LaTeX from sheet 'SCRNet'
\begin{table}[!t]
	\centering
	\caption{The impact of RGB color space correction and modulation parameters on model performance. ${\ell _{rgb}}$ denotes RGB color space, and ${\ell _{m}}$ denotes modulation parameters $({S_c}, {S_h})$.}
	\scalebox{0.92}{
	\begin{tabular}{l|cccccc}
		\toprule
		Model & ${\ell _{rgb}}$   & ${\ell _{m}}$ & NIQE↓  & BRISQUE↓ & PSNR↑  & SSIM↑ \\
		\midrule
		w/o JRS &       &       & 4.703  & 16.980  & 22.926  & 0.815  \\
		w/o ${\ell _{rgb}}$& \checkmark     &       & 4.341  & 15.737  & 23.635  & 0.830  \\
		w/o ${\ell _{m}}$&       & \checkmark     & 4.216  & 15.649  & 23.774  & 0.835  \\
		Ours  & \checkmark     & \checkmark     & \textcolor[rgb]{ 1,  0,  0}{4.039} & \textcolor[rgb]{ 1,  0,  0}{14.001} & \textcolor[rgb]{ 1,  0,  0}{24.901} & \textcolor[rgb]{ 1,  0,  0}{0.849} \\
		\bottomrule
	\end{tabular}%
}
	\label{tab:xr-rgb}%
\end{table}%

Fig. \ref{fig_us}(a) and Fig. \ref{fig_us}(b) show the average distribution of the BRISQUE and NIQE results on the 8 datasets for different methods. It can be seen from the figure that our proposed method achieved relatively better average distribution. In order to better demonstrate the quality of our enhancement results, we conduct a user study. Ten images are randomly selected from multiple datasets and enhanced using different methods. Twenty evaluators are invited to score the quality on a scale of 1 to 10, with 1 indicating extremely poor image quality and 10 indicating perfect enhancement quality. Importantly, these evaluations are blind (the evaluators do not know the corresponding enhancement method for the enhanced images).  Subsequently, the scores are summarized and averaged. For each method, the average score distribution given by the 20 evaluators is shown in Fig. \ref{fig_us}(c). From the figure, it can be seen that the quality scores for the URetinex method and the proposed JCRNet are between 5 and 9, indicating good user perception of the enhanced images.

\subsection{Ablation Study}

\begin{figure}[!t]
	\centering
	\includegraphics[width=0.48\textwidth]{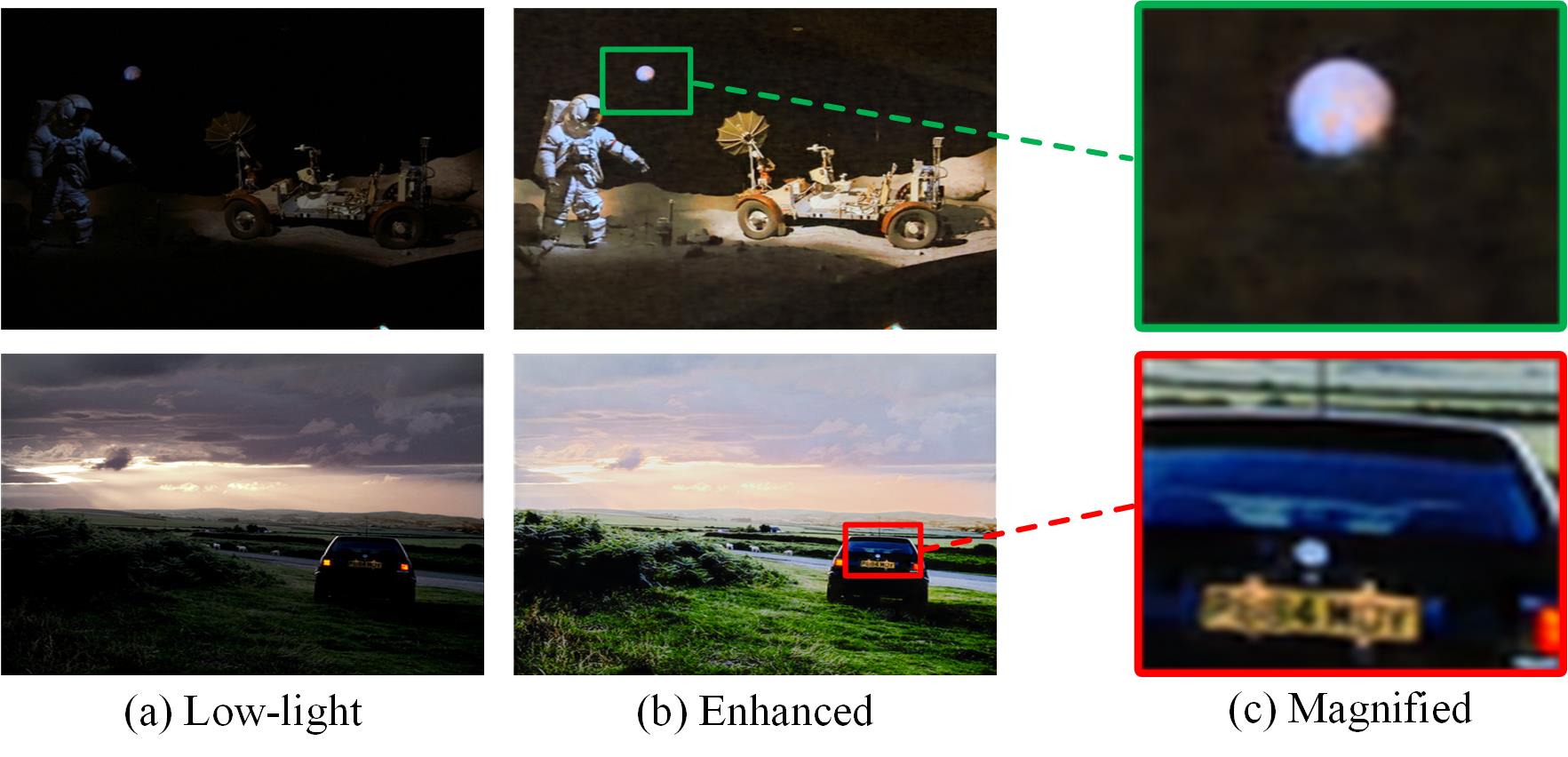}
	\DeclareGraphicsExtensions.
	\caption{Failure cases of proposed method.}
	\label{fig_fail}
\end{figure}

To analyze our method, we perform a series of ablation studies to demonstrate the contribution of each component of our model. We first verify the design of each stage in the proposed model by training models with different stages using the same strategy, and test them on six publicly available unreferenced datasets. The visual display is shown in Fig. \ref{fig_xr_stage}. 'Base' consists of a basic attention and an encoder-decoder. The quantitative results are shown in Table \ref{tab:xr-noref}. Secondly, on the LOL dataset, the validity of each component in the FES is verified. Finally, the contribution of the joint refinement stage in the model is validated using a low-light dataset under flash light.

\textbf{Design of the FES.} It is obvious from Fig. \ref{fig_xr_stage}(d) that the enhanced image in the model without FES has poor performance in some details, and the strong light area is slightly exposed. In addition, from the values of NIQE and BRISQUE reported in Table \ref{tab:xr-noref} and their mean values, it can also be found that removing FES affects the performance of the model. Therefore, FES is beneficial to the performance of the model.

\textbf{Design of the JRS.} Origin low-light image through the feature extraction stage, and then directly enters the illumination adjustment stage without passing the joint refinement stage. As can be seen from the enhancement results in Fig. \ref{fig_xr_stage}(c), the color saturation of the image obtained by removing JRS is low compared to the proposed method. This indicates that the ability of the model to adjust the details of the image, especially colors, becomes worse when the image is enhanced. In addition, the performance of the model has declined on the metrics reported in Table \ref{tab:xr-noref}. Therefore, the JRS in our proposed model has a certain refinement and adjustment ability, especially on the color level.

Specifically,  to more intuitively demonstrate the contribution of JRS, we test with low-light datasets under flash light specifically. As can be seen from Fig. \ref{fig_xr_jrs}(a), the color of the low-light dataset under flash light is slightly greenish, and the enhanced results of the proposed model are closer to the ground truth, indicating that the restored images have better visual effects. However, the enhancement results with removing JRS are shown in Fig. \ref{fig_xr_jrs}(b), and the image color is still somewhat greenish and visually poor. Furthermore, as evident from the quantitative evaluation metrics presented in Table \ref{tab:xr-fes}, it can be observed that the addition of JRS leads to improved quantitative metric scores, particularly in terms of naturalness metrics NIQE and BRISQUE, on the Toys dataset.

% Table generated by Excel2LaTeX from sheet 'SCRNet'
\begin{table}[!t]
	\centering
	\tabcolsep=0.13cm
	\caption{Quantitative comparison in terms of Parameters, FLOPs, and FPS. The higher the FPS value, the faster the test runs.}
	\scalebox{0.8}{
		\begin{tabular}{c|cccccc}
			\toprule
			Method & KinD \cite{zhang2019kindling}  & EnGAN \cite{jiang2021enlightengan} & URetinex \cite{wu2022uretinex} & FLW \cite{zhang2023fast}   & DDNet \cite{qu2023double} & Ours \\
			\midrule
			Params(M)↓ & 8.54  & 8.37  & 0.36  & 0.02  & 1.75  & 2.01  \\
			FLOPs (G)↓ & 36.57  & 72.61  & 233.09  & 2.47  & 66.31  & 219.74  \\
			FPS(s)↑ & 2.51  & 6.73  & 5.75  & 0.06  & 108.45  & 5.99  \\
			\bottomrule
		\end{tabular}%
	}
	\label{tab:comput}%
\end{table}%

To further visually demonstrate the impact of the JRS module on enhanced images, based on perceptual evaluations as mentioned in \cite{zhang2018unreasonable}, we create spatial distance transformation heatmaps between the enhanced images and the ground truth images, as shown in Fig. \ref{dis-hot}. Darker regions in the heatmaps indicate smaller differences from the ground truth images, and it can be observed from the figure that the enhanced results with the JRS module exhibit smaller spatial distance differences from the ground truth images. This indicates that after adding the JRS module for color correction and feature space transformation, the enhanced results retain more details such as color and texture. Additionally, ablation experiments are conducted on the modulation parameters for RGB color space correction and feature space transformation, and it can be concluded from the data reported that they both have a certain effect on improving the performance of the model. From the third and fourth rows of Table \ref{tab:xr-rgb}, it can be observed that the use of modulation parameters significantly improves the scores of PSNR and BRISQUE. This suggests that learnable modulation parameters progressively adjust during model training and encourage better learning of image details.

\textbf{Design of the IAS.} As shown in Fig. \ref{fig_xr_stage}(e), when origin low-light image through the feature extraction stage and the joint refinement stage, the enhanced image obtained without illumination adjustment is overexposed. In addition, as can be seen from the quantitative results reported in Table \ref{tab:xr-noref} and \ref{tab:xr-ref}, model performance has declined considerably. Therefore, the illumination adjustment stage is very important in the process of enhancement, which can solve the risk of overexposure that may occur in some images in the process of enhancement, improve the quality of images and maintain good visual effects.

% Table generated by Excel2LaTeX from sheet 'SCRNet'
\begin{table*}[!t]
	\centering
	\caption{Testing of real video frames captured by four different smartphones using different methods. Evaluation is done using the NIQE and BRISQUE assessment metrics. The proposed method performs the best on the average of all metrics. Best and second best scores are red and blue.}
	\begin{tabular}{l|cccccccccc}
		\toprule
		\multirow{2}[4]{*}{Methods} & \multicolumn{2}{c|}{Huawei Mate 20 Pro} & \multicolumn{2}{c|}{LG M322} & \multicolumn{2}{c|}{oppo r17} & \multicolumn{2}{c|}{Xiaomi Mi Mix 3} & \multicolumn{2}{c}{Total} \\
		\cmidrule{2-11}          & NIQE↓ & \multicolumn{1}{c|}{BRISQUE↓} & NIQE↓ & \multicolumn{1}{c|}{BRISQUE↓} & NIQE↓ & \multicolumn{1}{c|}{BRISQUE↓} & NIQE↓ & \multicolumn{1}{c|}{BRISQUE↓} & avg-NIQE↓ & avg-BRISQUE↓ \\
		\midrule
		CSDNet \cite{ma2021learning} & 5.587  & 60.416  & 5.632  & 52.905  & 5.651  & 56.504  & 6.225  & 57.552  & 5.774  & 56.844  \\
		IAT \cite{cui2022illumination}   & 5.556  & 62.492  & 6.313  & 76.081  & 5.841  & 62.635  & 7.413  & 74.281  & 6.281  & 68.872  \\
		KinD++ \cite{zhang2021beyond} & 5.818 & 50.300 & 5.457 & 47.167 & 5.481 & 57.929 & \textcolor[rgb]{ 0,  0,  1}{5.182} & \textcolor[rgb]{ 0,  0,  1}{56.891} & 5.484 & \textcolor[rgb]{ 0,  0,  1}{53.072}  \\
		LLFlow \cite{wang2022low} & 5.372 & 50.817 & 5.333 & \textcolor[rgb]{ 1,  0,  0}{52.641} & \textcolor[rgb]{ 1,  0,  0}{4.650} & 55.463 & 6.170 & 60.619 & \textcolor[rgb]{ 0,  0,  1}{5.381} & 54.885  \\
		RUAS \cite{liu2021retinex}  & 7.887  & 74.451  & 8.307  & 79.535  & 7.334  & 71.316  & 11.812  & 84.224  & 8.835  & 77.382  \\
		SCI \cite{ma2022toward}   & 7.258  & 70.047  & 7.626  & 83.738  & 6.525  & 64.355  & 10.822  & 86.350  & 8.058  & 76.123  \\
		SNR \cite{xu2022snr}  & 9.292  & 69.441  & 9.580  & 77.753  & 10.064  & 67.327  & 10.288  & 72.683  & 9.806  & 71.801  \\
		URetinex \cite{wu2022uretinex} & \textcolor[rgb]{ 1,  0,  0}{4.892} & 53.296 & 5.151 & 60.654 & 5.593 & \textcolor[rgb]{ 0,  0,  1}{53.362} & 5.952 & 69.365 & 5.397 & 59.169  \\
		Ours  & \textcolor[rgb]{ 0,  0,  1}{5.196} & \textcolor[rgb]{ 1,  0,  0}{49.946} & \textcolor[rgb]{ 1,  0,  0}{4.846} & \textcolor[rgb]{ 0,  0,  1}{53.139} & \textcolor[rgb]{ 0,  0,  1}{5.167} & \textcolor[rgb]{ 1,  0,  0}{51.321} & \textcolor[rgb]{ 1,  0,  0}{5.042} & \textcolor[rgb]{ 1,  0,  0}{50.381} & \textcolor[rgb]{ 1,  0,  0}{5.063} & \textcolor[rgb]{ 1,  0,  0}{51.197} \\
		\bottomrule
	\end{tabular}%
	\label{tab:flame}%
\end{table*}%

% Table generated by Excel2LaTeX from sheet 'SCRNet'
\begin{table*}[!t]
	\centering
	\caption{
		The low-light data (data1: RGB, data2: RGBD, data3: RGBT) are first enhanced using different enhancement methods, and then salient object detection experiments are conducted. Compared to others, our method's enhanced images were more effective in improving the performance of salient object detection tasks. Best and second best scores are red and blue.}
	\scalebox{0.81}{
		\begin{tabular}{l|cccccc|cccccc|cccccc}
			\toprule
			Datasets & \multicolumn{6}{c|}{Data1}                      & \multicolumn{6}{c|}{Data2}                    & \multicolumn{6}{c}{Data3} \\
			\midrule
			Methods & MAE↓  & maxF↑ & avgF↑ & wfm↑  & Sm↑   & Em↑   & MAE↓  & maxF↑ & avgF↑ & wfm↑  & Sm↑   & Em↑   & MAE↓  & maxF↑ & avgF↑ & wfm↑  & Sm↑   & Em↑ \\
			\midrule
			Low-light & 0.054 & \textcolor[rgb]{ 0,  0,  1}{0.936} & 0.904 & 0.883 & 0.888 & 0.906 & 0.045 & 0.920 & 0.900 & 0.870 & 0.904 & 0.923 & 0.056 & 0.821 & 0.793 & 0.743 & 0.833 & 0.892  \\
			CSDNet \cite{ma2021learning} & 0.048 & 0.933 & \textcolor[rgb]{ 0,  0,  1}{0.913} & 0.884 & 0.889 & 0.918 & 0.045 & 0.924 & 0.900 & 0.877 & 0.900 & \textcolor[rgb]{ 0,  0,  1}{0.938} & 0.052 & 0.858 & 0.818 & 0.779 & 0.856 & 0.899  \\
			SNR \cite{xu2022snr}   & 0.056 & 0.935 & 0.900 & 0.882 & 0.890 & 0.912 & 0.044 & 0.924 & 0.903 & 0.879 & 0.904 & 0.936 & \textcolor[rgb]{ 1,  0,  0}{0.043} & \textcolor[rgb]{ 0,  0,  1}{0.886} & \textcolor[rgb]{ 0,  0,  1}{0.842} & \textcolor[rgb]{ 0,  0,  1}{0.809} & \textcolor[rgb]{ 0,  0,  1}{0.872} & \textcolor[rgb]{ 0,  0,  1}{0.918}  \\
			URetinex \cite{wu2022uretinex} & 0.055 & 0.931 & 0.905 & 0.883 & \textcolor[rgb]{ 0,  0,  1}{0.893} & \textcolor[rgb]{ 0,  0,  1}{0.920} & 0.043 & 0.925 & 0.901 & 0.881 & 0.905 & \textcolor[rgb]{ 0,  0,  1}{0.938} & \textcolor[rgb]{ 0,  0,  1}{0.045} & 0.879 & 0.830 & 0.802 & 0.869 & 0.905  \\
			IAT \cite{cui2022illumination}   & \textcolor[rgb]{ 0,  0,  1}{0.046} & 0.933 & 0.907 & \textcolor[rgb]{ 0,  0,  1}{0.888} & 0.892 & 0.917 & \textcolor[rgb]{ 0,  0,  1}{0.041} & \textcolor[rgb]{ 0,  0,  1}{0.928} & \textcolor[rgb]{ 0,  0,  1}{0.907} & \textcolor[rgb]{ 0,  0,  1}{0.884} & \textcolor[rgb]{ 0,  0,  1}{0.907} & \textcolor[rgb]{ 1,  0,  0}{0.943} & \textcolor[rgb]{ 0,  0,  1}{0.046} & 0.882 & 0.840 & 0.806 & 0.869 & 0.909  \\
			Ours  & \textcolor[rgb]{ 1,  0,  0}{0.044} & \textcolor[rgb]{ 1,  0,  0}{0.940} & \textcolor[rgb]{ 1,  0,  0}{0.916} & \textcolor[rgb]{ 1,  0,  0}{0.895} & \textcolor[rgb]{ 1,  0,  0}{0.902} & \textcolor[rgb]{ 1,  0,  0}{0.929} & \textcolor[rgb]{ 1,  0,  0}{0.040} & \textcolor[rgb]{ 1,  0,  0}{0.931} & \textcolor[rgb]{ 1,  0,  0}{0.910} & \textcolor[rgb]{ 1,  0,  0}{0.888} & \textcolor[rgb]{ 1,  0,  0}{0.910} & \textcolor[rgb]{ 1,  0,  0}{0.943} & \textcolor[rgb]{ 1,  0,  0}{0.043} & \textcolor[rgb]{ 1,  0,  0}{0.900} & \textcolor[rgb]{ 1,  0,  0}{0.854} & \textcolor[rgb]{ 1,  0,  0}{0.823} & \textcolor[rgb]{ 1,  0,  0}{0.878} & \textcolor[rgb]{ 1,  0,  0}{0.923} \\
			\bottomrule
		\end{tabular}%
	}
	\label{tab:sod}%
\end{table*}%

\begin{figure}[!b]
	\centering
	\includegraphics[width=0.48\textwidth]{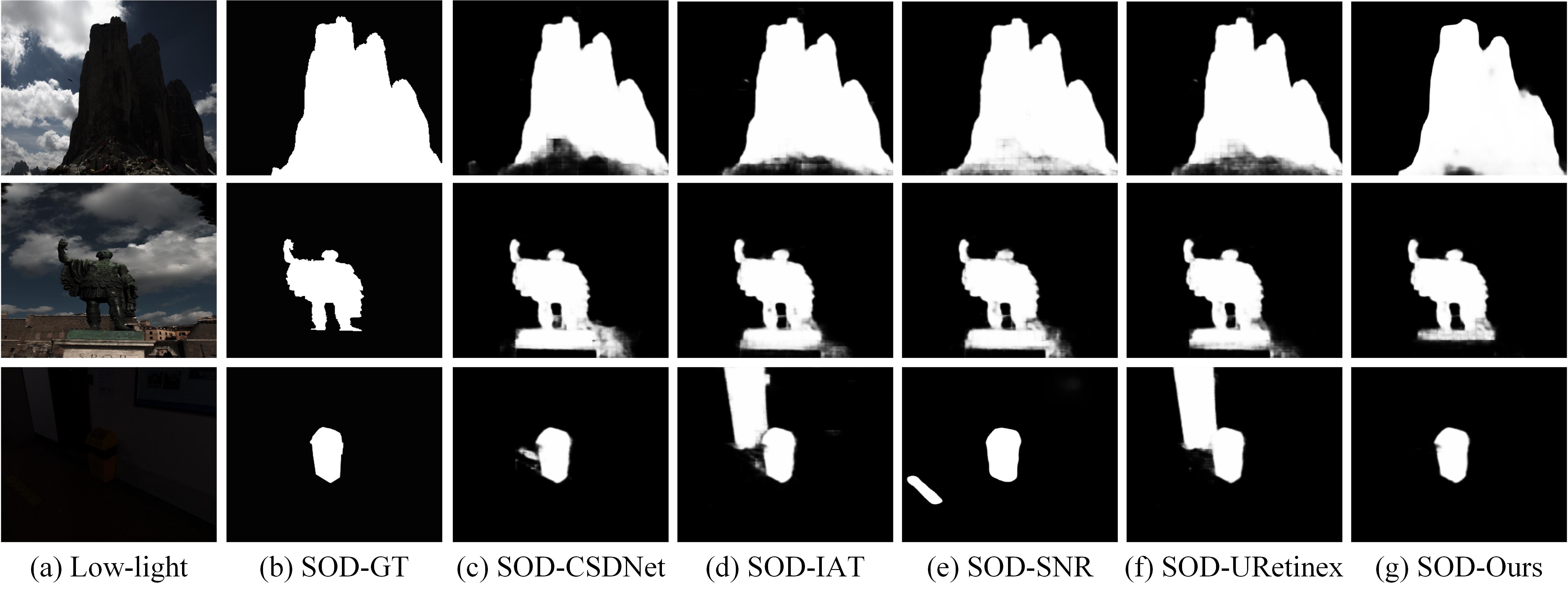}
	\DeclareGraphicsExtensions.
	\caption{Visual comparison results for saliency object detection (SOD) using enhanced images. (a) is a low-light image with salient objects. (b) is the ground truth of object detection. (c), (d), (e), (f), and (g) are the results of SOD using images enhanced by different methods.}
	\label{fig_sod}
\end{figure}

Especially, to highlight the contribution of IAS, Fig. \ref{fig_xr_jrs}(f) shows some enhanced results from the challenging ExDark dataset. It is obvious that the enhanced results without the use of illumination adjustment although the brightness is increased, the image is overexposed, resulting in an overall poor visual effect. However, the enhanced results with our proposed method are of moderate brightness, and there is no risk of overexposure.

The data distribution varies greatly among the 6 no-reference datasets, which makes them more challenging to work with. Additionally, we conduct ablation experiments on two reference-based datasets, LOL and COCO. From the data reported in Table \ref{tab:xr-ref}, it can be seen that different modules in the model have a significant impact on the reference-based datasets.

\textbf{Choices of components in the FES.} We test the contributions of the various components in the FES. Fig. \ref{fig_xr_componets} shows the visualized results by removing the different components. Moreover, to visualize the more obvious difference between the enhanced image and the ground truth, residual images are generated for each model enhanced image and ground truth respectively. The residuals between the enhanced results of JCRNet and the ground truth are minimal. In addition, quantitative evaluations of different models are reported in Table \ref{tab:xr-fes} (LOL dataset), PSNR and SSIM are significantly reduced when components are removed.

\subsection{Failure cases}

As shown in Fig. \ref{fig_fail}, although the proposed method achieves comprehensive performance improvements on most datasets, it can be challenging to achieve perfect enhancement in some extreme dark scenarios or in situations with minimal semantic information. For example, in the first row of the samples shown in Fig. \ref{fig_fail}, while the enhanced image improves details such as brightness and contrast, it introduces some unnecessary noise when dealing with purely black background areas. Therefore, more consideration of the data distribution is needed during the enhancement process. In the second row, the overall enhancement of the image is good, but for certain local areas, such as the enhancement of car windows, there seems to be some color deviations, making it difficult to achieve very satisfactory results when dealing with data that requires more semantic information.

\subsection{Computational Complexity}
Additionally, we analyze the computational complexity of the model. Table \ref{tab:comput} reports the model parameters, FLOPs, and average runtime (using the FPS metric) on 100 images of size 384×384. Smaller values for parameters and computational power indicate that the model requires fewer parameters and less computational power during runtime. A higher FPS indicates faster testing runtime. Compared to several representative and state-of-the-art models, the proposed model performs well in terms of parameters and runtime, although it does not show a significant advantage in computational power.

%\subsection{Analysis and Discussions}

\section{More applications} 

\textbf{Video frame enhancement in real-world scenarios.} To further validate the practicality of the proposed method in real videos \cite{liu2022instance}, we conduct experiments on video frames taken from four smartphones \cite{li2021low}: Huawei Mate 20 Pro (16 frames), Oppo R17 (26 frames), LG M322 (64 frames), and Xiaomi Mi Mix 3 (33 frames). We compare the results with the latest methods, as shown in Table \ref{tab:flame}. Since these data do not have paired ground truth images, we use two no-reference evaluation metrics for performance evaluation. Enhancing videos taken from smartphones is challenging due to the extreme darkness, motion blur, noise, and other artifacts present in the data. Therefore, most enhancement methods have limitations. Overall, our method achieved relatively better performance, as indicated by the reported values in the table.

\textbf{Validation of support for downstream visual tasks.} To further verify the practicality of the enhanced images in downstream tasks, we conduct a series of experiments on saliency detection tasks. Specifically, we first select low-light images from all datasets of RGB, RGB-depth (RGB-D) \cite{cheng2014depth} and RGB-thermal infrared (RGB-T) \cite{tu2022rgbt, wang2021cgfnet} for testing, with a total of 100 images of low-light data for RGB, 185 images of low-light data for RGB-D and 100 images for RGB-T. Based on the above data, we construct three low-light salient object detection datasets. Based on these three datasets, we use CSDNet, SNR, URetinex, IAT, and our proposed method to enhance the low-light RGB images. Then, the original low-light images and the images enhanced by different methods are input into the same saliency object detection model \cite{liu2021visual} based on a Transformer architecture for saliency object detection tasks. Finally, the detection results are visualized, as shown in Fig. \ref{fig_sod}. It can be seen that the proposed method’s enhanced images obtain relatively more accurate saliency objects in the detection task. Additionally, we evaluate detection performance using five commonly used quantitative evaluation metrics in saliency object detection tasks, as shown in Table \ref{tab:sod}. The proposed method uses joint refinement and dynamic illumination adjustment to effectively control the balance of brightness, color, and exposure, making it less likely to introduce bias during saliency detection. From the reported metrics in Table \ref{tab:sod}, it can be seen that our method’s enhanced images are more conducive to improving saliency detection performance compared to several of the latest enhancement methods.

\section{Conclusion}

In this study, we introduce a novel architecture for improving low-light images. Our approach consists of three key stages. First, we extract global and local image features using residual blocks, residual channel attention blocks, and encoder-decoder blocks. Next, we incorporate a self-supervised block to generate supervised feature mappings, effectively guiding the image enhancement process. Most importantly, we address color distortion issues during the enhancement process through a joint refinement stage. We utilize residual maps generated in lightening and darkening blocks to fine-tune exposure balance in the illumination adjustment stage. Extensive experiments demonstrate that our proposed JCRNet outperforms existing models in resolving color distortion and exposure imbalances in low-light images. Furthermore, our model proves its effectiveness across different types of low-light images and showcases practicality in saliency detection tasks.

In the future, we aim to optimize our model for more complex low-light enhancement tasks in diverse environments and explore ways to perform such enhancements efficiently with limited computational resources.

%\section*{Acknowledgment}
%This research was supported by the National Natural Science Foundation of China (61772319, 62002200, 61976125, 61976124), Shandong Natural Science Foundation of China (ZR2017MF049).

\footnotesize{
\bibliographystyle{IEEEtran}
\bibliography{mybibfile.bib}
}
\ifCLASSOPTIONcaptionsoff
  \newpage
\fi

\end{document}